\documentclass[letterpaper]{article} \usepackage[preprint]{aaai2027}

\usepackage[hyphens]{url} \usepackage{graphicx}   \usepackage{natbib} \usepackage{caption} 

\usepackage{amsfonts}
\usepackage{amsmath}
\usepackage{amssymb}
\usepackage{mathtools}
\usepackage{amsthm}
\usepackage{booktabs}
\usepackage{makecell}
\usepackage{multirow}
\usepackage{xcolor}
\usepackage{algorithm}
\usepackage{algpseudocode}

\usepackage{stringenc}

\StringEncodingConvert{\PDFAuthorUTF}{Björn Ludwig, Alexander Windmann, René Heesch, Oliver Niggemann}{utf8}{utf16be}
\EdefEscapeHex{\PDFAuthorHex}{\PDFAuthorUTF}

\theoremstyle{plain}
\newtheorem{theorem}{Theorem}[section]
\newtheorem{proposition}[theorem]{Proposition}

\theoremstyle{definition}

\theoremstyle{remark}

\title{From Reconstruction to Imputation: \\ Test-Time Optimization for Missing Time-Series Channels}
\author{Björn Ludwig\corresponding, Alexander Windmann, René Heesch, Oliver Niggemann}
\affiliations{
    HSU-AI -- Institute for Artificial Intelligence\\
    Helmut Schmidt University, Hamburg, Germany\\
    bjoern.ludwig@hsu-hh.de
}

\begin{document}

\maketitle

\begin{abstract}
Complete sensor-channel failures create a difficult form of time-series missingness: local interpolation is unavailable, and deploying a separately trained imputer for every failure pattern may be impractical. We ask whether a sequence reconstructor trained without artificial mask augmentation and then frozen can be converted into an imputer at inference time. Our method treats the missing entries of each test window as optimization variables, clamps all observations, and backpropagates through the frozen model to minimize reconstruction error only on observed coordinates. The resulting observed-self-consistency objective requires neither artificial mask augmentation nor full-channel dropout during reconstructor training. We characterize when this refinement is informative through the cross-Jacobian from missing inputs to observed residuals, and use an optional temporal-smoothness prior when that coupling is weak. Across five real-world datasets with exactly one channel synthetically masked per test window and four reconstruction architectures, no-prior test-time refinement reduces MAE relative to one-pass reconstruction in 19 of 20 dataset–backbone combinations. Compared with mask-aware imputers trained under either standard structured masks or matched full-channel dropout, the best refined reconstructor achieves the lowest MAE on two datasets and trails the strongest baseline on three. These gains require substantially more inference time and are sensitive to cross-channel coupling. Our results establish test-time self-consistency as a conditional inference-time adapter for frozen reconstruction models that complements mask-aware imputation without retraining the reconstructor.
\end{abstract}

\section{Introduction}
\label{sec:introduction}

Missing values are common in multivariate time series from environmental monitoring, energy systems, and traffic sensing \cite{de2008field_Italy_air,zhang2017cautionary_Beijing_air,electricityloaddiagrams20112014_321,liu2023largest}. An especially hard case arises when one complete channel is unavailable. Unlike isolated points or short gaps, the missing channel provides no local observations within the affected window, so recovery must rely on temporal structure learned elsewhere and on dependencies with the remaining channels \cite{Khayati_Mind_the_gap,cini2022filling,mitra2023learningdatastructuredmissingness}.

\begin{figure}[h]
    \centering
    \includegraphics[width=1\linewidth]{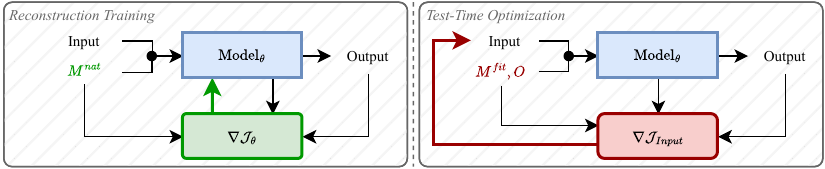}
    \caption{\textbf{High-level overview.} Left: train a reconstructor on baseline-filled windows using observed-entry loss. Right: TTO updates missing inputs using frozen model to reduce fit-observed residuals and yield the imputation.}
    \label{fig:HighLevelMethod}
\end{figure}

This paper asks a specific operational question: \emph{Given a frozen sequence reconstructor, can it be turned into an imputer for a previously unseen full-channel mask without retraining its weights?}

Deep recurrent, attention-based, and diffusion-based imputers learn conditional completions or sampling procedures during training \cite{BRITS_cao2018britsbidirectionalrecurrentimputation,SAITS_Du_2023,tashiro2021csdi}. Mask-aware methods can explicitly condition on the observation mask and enforce consistency with available values. Instead of learning or amortizing the completion for the target mask, we optimize the missing coordinates of each test window by differentiating through a fixed reconstructor. A useful reconstructor must couple the missing coordinates to observed outputs through a bottleneck or another learned dependency, since an identity-like model whose observed outputs are independent of the missing inputs provides no gradient with which to recover the missing channel.

Figure~\ref{fig:HighLevelMethod} summarizes the two phases. We first train a sequence reconstructor with a reconstruction loss on observed entries and without artificial mask augmentation, then freeze its parameters. At inference time, Test-Time Optimization (TTO) treats missing entries as free variables, clamps observations, and minimizes the frozen model residual only on observed coordinates. The resulting observed-self-consistency objective can incorporate simple priors or feasibility constraints when the data term alone is underconstrained. Its informativeness is governed locally by the cross-Jacobian from optimized missing inputs to residuals on observed outputs: if this coupling is weak or rank-deficient, multiple completions can be similarly self-consistent.

\paragraph{Contributions.}
We make four contributions. (1) We formulate observed-self-consistency TTO as an inference-time adapter that turns a frozen differentiable reconstructor into a point-estimate imputer while clamping observations and fixing model parameters. (2) We characterize its local informativeness through the missing-to-observed cross-Jacobian and explain how temporal smoothness regularizes weakly constrained directions. (3) Across five real-world datasets, four architectures, and synthetic one-channel masks, TTO improves over forward-only reconstruction in 19 of 20 dataset--backbone combinations. (4) Against mask-aware imputers trained with structured masks or matched full-channel dropout, the best TTO result wins on ETTh1 and Electricity but trails on BeijingAir, ItalyAir, and PeMS, while requiring substantially more inference time.

Together, these results support TTO as a conditional adapter when a frozen reconstructor provides informative missing-to-observed coupling, not as a blanket replacement for mask-aware imputation models.
\section{Related Work}
\label{sec:related_work}

\paragraph{Deep time series imputation: from recurrent baselines to modern generative models.}
Deep imputation methods have become the dominant approach for handling missing values in multivariate time series, as they can model nonlinear dynamics and cross-variable dependencies under structured missingness \cite{wangDeepLearningMultivariate2025}.
Most deep imputers perform amortized inference, producing imputations by evaluating a trained network (or a fixed learned sampling procedure) on a partially observed window, rather than optimizing missing values per instance \cite{BRITS_cao2018britsbidirectionalrecurrentimputation, Mattei2019MIWAEDG, tashiro2021csdi, SAITS_Du_2023}.
Within this paradigm, the community has progressed from recurrent baselines to attention-based and generative formulations, with each generation improving robustness to longer-range dependencies and more complex missingness patterns \cite{wen2023transformerstimeseriessurvey}.

\paragraph{Predictive (point-estimate) imputers.}
Recurrent models constitute a canonical starting point due to their sequential inductive bias.
Earlier RNN-based approaches such as GRU-D introduced explicit decay mechanisms to handle irregular sampling and missingness, influencing later recurrent imputers \cite{GruD_che2016recurrentneuralnetworksmultivariate}.
BRITS is a widely used recurrent baseline that performs bidirectional recurrent imputation to propagate information from both temporal directions \cite{BRITS_cao2018britsbidirectionalrecurrentimputation}.
Beyond recurrence, attention mechanisms improved long-range modeling under missingness; SAITS is a representative attention-based imputer that leverages masked self-attention blocks and missingness indicators \cite{SAITS_Du_2023}.

Graph-based formulations explicitly model cross-variable structure by learning dependencies between features.
GRAPE is a prominent example that addresses missingness through graph representation learning, making cross-channel coupling an explicit modeling object rather than an implicit byproduct of sequence modeling \cite{you2020handling}.
In parallel, convolutional and frequency-inspired designs demonstrate that strong performance can be achieved by restructuring temporal variation.
TimesNet is a competitive backbone that uses FFT-based reformulation and is frequently adopted as a high-performing baseline across time-series tasks \cite{wu2023timesnettemporal2dvariationmodeling}.

\paragraph{Generative imputers and uncertainty.}
Generative approaches address the ambiguity of imputation by learning distributions over plausible completions rather than producing a single point estimate. MIWAE provides a likelihood-based framework for imputing incomplete data with deep latent-variable models \cite{Mattei2019MIWAEDG}, while GP-VAE incorporates a Gaussian-process prior to model temporal dependencies under missingness \cite{fortuin2020gpvaedeepprobabilistictime}.
Diffusion and score-based methods are an important generative paradigm for time-series imputation; CSDI is a representative diffusion-based method that conditions generation on observed values to impute missing segments \cite{tashiro2021csdi}.
Overall, these generative methods provide principled ways to represent uncertainty. Although diffusion samplers are themselves iterative, their test-time iterations follow a learned conditional denoising process rather than directly optimizing missing entries for each instance through a frozen reconstructor.

\paragraph{Iterative imputation and test-time refinement.}
A complementary line of work emphasizes \emph{iterative} imputation, where solutions are refined through repeated updates rather than a single feed-forward pass.
HyperImpute formalizes generalized iterative imputation with automatic model selection \cite{jarrett2022hyperimpute}, and IGRM performs imputation through iterative graph reconstruction \cite{Zhong_2023_data_imputation_iterative_graph}. Related perspectives include distribution-matching objectives for missing-value imputation \cite{zhao2023transformeddistributionmatchingmissing}.
Our method aligns with the broader message that per-instance refinement can be beneficial, but differs in mechanism and scope: instead of alternating model fitting and filling or employing bespoke iterative update rules, we optimize missing entries \emph{directly} by differentiating through a \emph{frozen} sequence reconstructor, yielding an inference-time adapter for differentiable reconstructors with informative missing-to-observed coupling.
Conceptually, our method follows the general idea of optimizing inputs while keeping a differentiable model fixed, as in differentiable planning \cite{wu2020scalable_planning} and network inversion \cite{haim2022reconstructing}. Unlike these settings, our objective is mask-conditioned imputation as we optimize only missing entries to enforce self-consistency on observed values, rather than maximizing reward or reconstructing training data.

\paragraph{Takeaway.}
Taken together, current competitive imputers provide powerful amortized mappings across diverse architectures, but complete-channel masks remain difficult because no within-channel observations are available and recovery depends on learned cross-channel structure.
This motivates our focus on test-time optimization: enforce observed self-consistency under a frozen reconstructor and regularize ill-posed cases with simple priors and feasibility constraints, conditionally bridging amortized imputation with iterative per-instance inference rather than replacing mask-aware models.
\section{Problem Setting}
\label{sec:problem_setup}

We consider a window $X\in\mathbb{R}^{T\times D}$ with $D$ measurement channels and length $T$ . We use \emph{channel} for one dimension $d\in\{1,\ldots,D\}$ and \emph{coordinate} for one time--channel entry $(t,d)$.

\paragraph{Mask contract.}
Let $M^{\mathrm{nat}}\in\{0,1\}^{T\times D}$ be the native-observed mask, where $M^{\mathrm{nat}}_{t,d}=1$ exactly when $X_{t,d}$ is available before controlled evaluation masking. To evaluate complete-channel recovery, we choose one channel $d^\star$ and define the evaluation-missing mask
\begin{equation}
M^{\mathrm{eval}}_{t,d}:=\mathbb{I}[d=d^\star],
\label{eq:full_channel_mask}
\end{equation}
so one entire channel is synthetically held out in each test window. The selected channel is balanced across the test set. We then define
\begin{equation}
\begin{aligned}
M^{\mathrm{fit}}&:=M^{\mathrm{nat}}\odot(1-M^{\mathrm{eval}}),\\
O&:=1-M^{\mathrm{fit}},\\
H&:=M^{\mathrm{nat}}\odot M^{\mathrm{eval}}.
\end{aligned}
\label{eq:mask_contract}
\end{equation}
Here $M^{\mathrm{fit}}$ marks entries available to the observed-residual objective, $O$ marks coordinates optimized by TTO, and $H$ marks artificial holdouts that have reference values and can therefore be scored. Thus $O$ includes both native missing entries and artificial holdouts, whereas $H$ excludes native missing entries. We refer to $M^{\mathrm{nat}}$, $M^{\mathrm{eval}}$, $M^{\mathrm{fit}}$, and $O$ as the native-observed, evaluation-missing, fit-residual, and optimization masks, respectively.

\paragraph{Conversion contract.}
The inference-time adapter receives $X$, $M^{\mathrm{fit}}$, $O$, and a frozen reconstructor $f_\theta$. It keeps $\theta$ fixed, clamps entries selected by $M^{\mathrm{fit}}$, and changes only candidate values on $O$. If $\bar X=F_\theta(X,M^{\mathrm{fit}})$ denotes the reconstructor prediction after optimization, the completed window is
\begin{equation}
\hat X:=M^{\mathrm{fit}}\odot X+O\odot\bar X.
\label{eq:completed_series}
\end{equation}
Consequently, every value available to the adapter remains unchanged in the completed output.

\paragraph{Evaluator-only targets.}
For controlled evaluation, $\tilde X$ denotes reference values on the artificial holdouts $H$. Neither $\tilde X$ nor $H$ is supplied to TTO as the adapter sees only the combined availability/optimization masks $M^{\mathrm{fit}}$ and $O$. Native missing entries are not scored. The reported mean absolute error is
\begin{equation}
\mathcal{E}_{\mathrm{MAE}}(\bar X;\tilde X,H)
:=\frac{\|H\odot(\bar X-\tilde X)\|_1}{\|H\|_1}.
\label{eq:eval-objective}
\end{equation}
At deployment, $M^{\mathrm{fit}}$ can be supplied directly from the observed/missing pattern, the decomposition into $M^{\mathrm{nat}}$, $M^{\mathrm{eval}}$, and $H$ is evaluation bookkeeping only.
\section{Method}
\label{sec:method}

\begin{figure*}[h]
    \centering
    \includegraphics[width=1\linewidth]{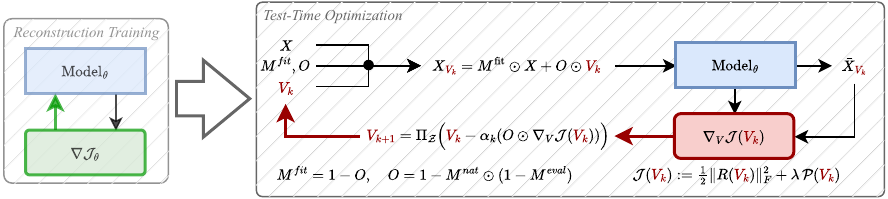}
    \caption{\textbf{Converting a Frozen Reconstructor into an Imputer.} In the prior/offline phase, an autoencoder learns to reconstruct observed entries from baseline-filled windows; the loss is evaluated only where training data are natively observed. In the proposed post-training conversion, the reconstructor is frozen and candidate missing values $V$ are optimized through it to minimize the fit-observed reconstruction residual (optionally regularized). Observed inputs remain clamped, and gradients update only optimized entries.}
    \label{fig:MethodDetails}
\end{figure*}

As illustrated in Fig.~\ref{fig:MethodDetails}, the method converts a trained sequence reconstructor into an imputer without updating its parameters or accessing held-out values. It consists of two phases: offline reconstruction training supplies a learned sequence prior, after which post-training test-time optimization refines only the missing entries by enforcing observed self-consistency. The following subsections specify these phases and the resulting optimization procedure.

\subsection{Phase 1: Offline Reconstruction Training}
\label{sec:method:model}
We train a standard autoencoder as a sequence reconstructor by reconstructing observed entries of baseline-filled windows.
Let $(X,M^{\mathrm{nat}})$ denote a training window and its native-observed mask (Sec.~\ref{sec:problem_setup}), and let $B(\cdot)$ be a baseline fill operator (e.g., zero or per-channel mean). The reconstructor is an encoder--decoder trained from
\begin{equation}
\begin{aligned}
X^{\mathrm{in}}&:=M^{\mathrm{nat}}\odot X + (1-M^{\mathrm{nat}})\odot B(X),\\
X^{\mathrm{rec}}&:=f_\theta(X^{\mathrm{in}}).
\end{aligned}
\end{equation}
by minimizing the reconstruction loss on native-observed entries,
\begin{equation}
\min_\theta\ \mathbb{E}_{(X,M^{\mathrm{nat}})}\Big[\ \|M^{\mathrm{nat}}\odot(f_\theta(X^{\mathrm{in}})-X)\|_F^2\ \Big].
\label{eq:method:ae_train}
\end{equation}
The mask $M^{\mathrm{nat}}$ affects baseline filling and loss masking but is not provided as a separate network input, and training uses no artificial masking/denoising objective. After training, the reconstructor is frozen so the conversion stage below changes only missing input coordinates and does not retrain $\theta$.

\subsection{Phase 2: Post-Training Conversion by Input Optimization}
\label{sec:method:tti}
At test time, we use the shared mask contract defined in section \ref{sec:problem_setup}. The adapter receives only $X$, the fit-residual mask $M^{\mathrm{fit}}$, and the optimization mask $O$. It does not receive the evaluator-only target mask $H$ or the reference values $\tilde X$. We freeze $\theta$ and optimize only coordinates selected by $O$.
We introduce an explicit optimization variable $V\in\mathbb{R}^{T\times D}$ storing candidate values at optimized coordinates and define the completed input
\begin{equation}
X_V := M^{\mathrm{fit}}\odot X + O\odot V,
\qquad
\bar X_V := f_\theta(X_V).
\label{eq:method:defs_V}
\end{equation}
By construction, $X_V$ agrees with $X$ on observed indices, thus $f_\theta(X_V)$ already provides a one-shot completion, which we further refine by optimizing $V$.

\paragraph{From reconstructor to imputation model.}
Under the conversion contract in Sec.~\ref{sec:problem_setup}, our imputation model composes a learned reconstructor with a test-time optimization over missing entries.
We choose $V$ by enforcing \emph{observed self-consistency} via the residual $R(V) := M^{\mathrm{fit}}\odot(\bar X_V - X)$ and solve
\begin{equation}
\begin{aligned}
V^* &\in \operatorname*{arg\min}_{V\in\mathcal Z}\ \mathcal{J}(V),
 \text{ with } \\
 \mathcal{J}(V) &:= \tfrac12\|R(V)\|_F^2 + \lambda\,\mathcal{P}(V),
\end{aligned}
\label{eq:method:objective}
\end{equation}
where $\mathcal Z$ encodes optional constraints (e.g., value bounds),
$\mathcal{P}(V)$ is an optional prior on missing trajectories (e.g., temporal smoothness or a proximal term),
and $\lambda\ge 0$ controls the prior strength (specified in Sec.~\ref{sec:experiments}). An informative $\mathcal{P}(V)$ can further improve imputation when domain knowledge is available, but the method remains well-defined without it (i.e., $\lambda=0$).
Crucially, the data term depends only on $M^{\mathrm{fit}}$ to ensure neither evaluator-only reference values on $H$ nor baseline-filled native missing entries provide supervision.
The resulting imputation output is
\begin{equation}
\bar X = F_\theta(X,M^{\mathrm{fit}}) := \bar X_{V^*} = f_\theta(X_{V^*}),
\end{equation}
and the completed series is \text{\ensuremath{\hat X := M^{\mathrm{fit}}\odot X + O\odot \bar X}} as in Eq.~\eqref{eq:completed_series}.
We return the reconstructor output $f_\theta(X_{V^*})$, rather than the auxiliary variable $V^*$ itself, because $V^*$ is optimized only to elicit an observed-self-consistent response from the frozen model. The output is the model-consistent completion learned by the reconstruction objective and keeps forward-only and optimized inference comparable. Directly returning $V^*$ would bypass the learned reconstruction mapping.
We next discuss initialization and when the resulting refinement problem is locally well-posed.

\paragraph{Initialization.}
We initialize the optimization variable with a baseline completion,
\begin{equation}
V_0 := B(X),
\label{eq:method:init}
\end{equation}
which yields the initial completed input $X_{V_0}$ and prediction $\bar X_{V_0}$. In our experiments, we use per-channel mean of the training set.

\paragraph{Well-posedness of the refinement problem.}
The refinement step is informative only if changing missing entries affects the model's predictions on observed entries, i.e., if the observed residual $R(V)=M^{\mathrm{fit}}\odot(f_\theta(X_V)-X)$ depends on $V$ through the optimized coordinates.
Let $\Omega=\{(t,d):M^{\mathrm{fit}}_{t,d}=1\}$ be the fit-observed set and $\bar\Omega=\{(t,d):O_{t,d}=1\}$ the optimized set. Let $r(V)\in\mathbb{R}^{|\Omega|}$ stack $R(V)$ on $\Omega$, and let $V_{\bar{\Omega}}\in\mathbb{R}^{|\bar{\Omega}|}$ collect optimized entries.
We define the \emph{cross-Jacobian block}
\begin{equation}
C(V)\;:=\;\frac{\partial r(V)}{\partial V_{\bar{\Omega}}}
\ \in\ \mathbb{R}^{|\Omega|\times|\bar{\Omega}|},
\label{eq:method:cross_jac_def}
\end{equation}
i.e., the sensitivity of the observed residual to perturbations of the missing entries.
Note that $R(V)$ depends on $V$ only through optimized coordinates because $X_V$ clamps entries in $M^{\mathrm{fit}}$.

This coupling condition is necessary for an informative refinement signal. For example, if $f_\theta$ is the identity map, its observed outputs depend only on the clamped observed inputs. Consequently, if $C(V)=0$ the data-term gradient provides no information about the missing coordinates.

Under the differentiability and exact-self-consistency assumptions in Appendix~\ref{app:cross_jac}, if $C(V^\star)$ has full column rank at a point $V^\star$ with $R(V^\star)=0$, then $V^\star$ is isolated among nearby self-consistent completions and the local Gauss--Newton curvature is governed by $C(V^\star)^\top C(V^\star)$.
This is a local statement around an exact solution and does not establish global identifiability or rule out distant alternative completions.
If $C(V^\star)$ is rank-deficient or ill-conditioned, multiple nearby completions can explain the observed entries equally well, making the refinement sensitive.
The optional prior $\lambda\,\mathcal{P}(V)$ regularizes directions that observed self-consistency leaves weakly constrained. It can stabilize or select among such solutions, but it is not the source of the basic conversion (Appendix~\ref{app:cross_jac}).

\subsection{Optimization algorithm}
\label{sec:method:opt}
We minimize Eq.~\eqref{eq:method:objective} by projected first-order methods, backpropagating through the frozen reconstructor $f_\theta$.
Let $G_k := \nabla_V \mathcal{J}(V_k)$ computed by automatic differentiation.
In the first step we update only missing coordinates by using a defined step size $\alpha_k$ and in the second step we (optionally) use a projection $\Pi_{\mathcal Z}$ to $\mathcal Z$:
\begin{align}
V_{k+\tfrac12} &= V_k - \alpha_k\big(O\odot G_k\big),\\
V_{k+1} &= \Pi_{\mathcal Z}(V_{k+\tfrac12}),
\end{align}
with early stopping based on the observed residual $\|R(V_k)\|_F$ or relative objective decrease.
Algorithm~\ref{alg:tti_short_R} summarizes the resulting test-time optimization procedure.

\begin{algorithm}[t]
\caption{\textbf{T}est-\textbf{T}ime \textbf{O}ptimization (TTO) of missing values (frozen reconstructor)}
\label{alg:tti_short_R}
\begin{algorithmic}[1]
\Require input window $X$, masks $M^{\mathrm{fit}},O$, frozen reconstructor $f_\theta$, prior weight $\lambda$, steps $K$
\State Initialize $V \gets B(X)$ \Comment{baseline fill for missing entries}
\For{$k=1,\dots,K$}
  \State $X_V \gets M^{\mathrm{fit}}\odot X + O\odot V$ \Comment{clamp fit entries}
  \State $\bar X_V \gets f_\theta(X_V)$
  \State $R(V) \gets M^{\mathrm{fit}}\odot(\bar X_V - X)$ \Comment{observed residual}
  \State $\mathcal{J} \gets \tfrac12\|R(V)\|_F^2 + \lambda\,\mathcal{P}(V)$
  \State $G \gets \nabla_V \mathcal{J}$ \Comment{backprop through $f_\theta$; weights frozen}
  \State $V \gets \Pi_{\mathcal Z}\!\bigl(\textsc{OptStep}(V,O\odot G)\bigr)$ \Comment{e.g., Adam; optional constraints}
  \If{\textsc{Converged}($\mathcal{J}$)}
    \State \textbf{break}
  \EndIf
\EndFor
\State \Return $\hat X \gets M^{\mathrm{fit}}\odot X + O\odot f_\theta(M^{\mathrm{fit}}\odot X + O\odot V)$
\end{algorithmic}
\end{algorithm}
In our experiments, the optional $\mathcal{P}(V)$ is instantiated as a first-order temporal smoothness penalty on missing indices.
Full details and stopping criteria are given in Appendix~\ref{appendix-Algorithm}.

\paragraph{Runtime.}
Test-time optimization requires multiple forward/backward passes per window, which we implement in a batched, fully vectorized manner. Implementation details are provided in Appendix~\ref{appendix-Algorithm} and the full runtime comparison is stated in Appendix~\ref{app:clock_runtime}.
\section{Experiments}
\label{sec:experiments}

Our main goal is to evaluate the effectiveness and trade-offs of \emph{test-time optimization} (Sec.~\ref{sec:method:tti}) in \emph{hard missingness regimes} that stress cross-coordinate coupling.
In particular, we use a synthetic evaluation mask that removes exactly one complete channel from each test window.
Throughout, we report $\mathcal{E}_{\mathrm{MAE}}(\bar X;\tilde X,H)$ from Eq.~\eqref{eq:eval-objective}. The scoring mask $H$ contains only natively observed entries in the synthetically held-out channel, so native missing values never enter the reported errors.

\subsection{Evaluation Questions}
\label{sec:experiments:rqs}
We organize the evaluation around three questions addressing effectiveness, comparison, and optimization behavior and cost: \textbf{(EQ1)} Under an unseen full-channel mask, does TTO improve a frozen reconstructor over forward-only completion? \textbf{(EQ2)} How does the converted reconstructor compare with mask-aware imputers trained using either standard structured masks or matched full-channel masks? \textbf{(EQ3)} What optimization behavior and inference-time cost does conversion exhibit?

\begin{figure*}[t]
    \centering
    \includegraphics[width=\textwidth]{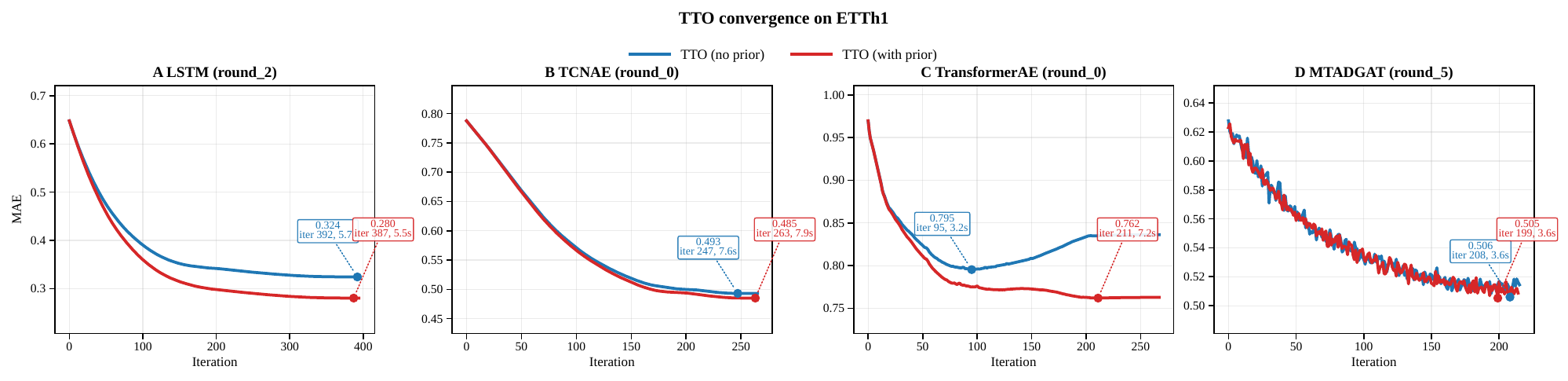}
    \caption{\textbf{Representative TTO convergence on ETTh1.} Imputation MAE over optimization iterations for four frozen reconstruction backbones, with (red) and without (blue) the smoothness prior. The trajectories are smooth in this representative case and stabilize within a small fraction of the conservative $K=5000$ cap, using objective-based early stopping.}
    \label{fig:TTOconvergence}
\end{figure*}

\subsection{Datasets and preprocessing}
\label{sec:experiments:data}
We evaluate on five multivariate time series from different domains:
\textit{BeijingAir} (air quality), \textit{ItalyAir} (air quality), \textit{ETTh1} (energy/temperature),
\textit{Electricity} (electricity load), and \textit{PeMS} (traffic).
We follow the standard time-based splitting and windowing protocol used in TSI-Bench \cite{du2024tsibenchbenchmarkingtimeseries} to avoid leakage.
All datasets are standardized using statistics from the training split. For full details, see \citet{du2024tsibenchbenchmarkingtimeseries}. In our experiments only the \textit{BeijingAir} Dataset has a low rate of native missingness of $1.6\%$.

\subsection{Missingness patterns and evaluation settings}
\label{sec:experiments:missingness}

\paragraph{Structured missingness for deep imputation baselines.}
Following the TSI-Bench protocol \cite{du2024tsibenchbenchmarkingtimeseries}, we evaluate deep imputation baselines under three structured missingness patterns:
\emph{point}, \emph{subsequence}, and \emph{block} missingness, each at a 10\% missing rate.
For these baselines, we adopt the standard practice of training and evaluating under the same pattern family
(point/subsequence/block) to match the intended operating regime of mask-aware imputers.

\paragraph{Full-channel missing evaluation.}
We use the mask construction in Sec.~\ref{sec:problem_setup}. Thus, for each test window, one channel $d^\star$ is selected and $M^{\mathrm{eval}}_{t,d}=\mathbb{I}[d=d^\star]$ for every time step. The fit-residual and optimization masks then follow from Eq.~\eqref{eq:mask_contract}. We assign $d^\star$ cyclically so each channel is held out equally often across the test set.

\paragraph{Training-mask regimes for baseline comparison.}
Table~\ref{tab:full_missing_test} summarizes mask-aware baselines with two per-dataset lower envelopes: \emph{Best OOD mask-aware} over checkpoints trained with 10\% point, subsequence, or block corruption and evaluated under full-channel corruption, and \emph{Best matched mask-aware} over checkpoints trained with matched full-channel dropout.
Appendix~\ref{appendix-experiments-full-channel-detailed} reports every constituent checkpoint with its \emph{point}, \emph{subseq}, \emph{block}, or \emph{full feature} label.
In contrast, we train and tune the reconstruction architectures for this study using only native missingness, without full-channel masks or additional artificial masking. We then freeze their weights and apply TTO in Eq.~\eqref{eq:method:objective}.
The comparison therefore has a deliberate supervision boundary: mask-aware models receive explicit corruption training, whereas reconstructors do not.

\subsection{Compared methods.}
\label{sec:experiments:baselines}
We compare against strong baselines covering major architecture families:
BRITS (RNN) \cite{BRITS_cao2018britsbidirectionalrecurrentimputation}, SAITS (masked self-attention) \cite{SAITS_Du_2023}, iTransformer (Transformer) \cite{liu2024itransformerinvertedtransformerseffective}, TimesNet (CNN/FFT-style) \cite{wu2023timesnettemporal2dvariationmodeling}, and
FreTS (MLP/spectral) \cite{yi2023frequencydomain}.
Several of the considered backbones were originally developed for forecasting.
We use them here as \emph{representative architectures} by following the standard adaptation procedure in the imputation literature (notably as operationalized by \citet{SAITS_Du_2023}).
Model configurations for all baselines follow the fixed hyperparameter settings released with TSI-Bench \cite{du2024tsibenchbenchmarkingtimeseries}.
As a non-neural sanity check, we additionally report \textsc{MeanFill}, which replaces missing entries by per-channel means from the corresponding training split.

\subsection{Our models and hyperparameters.}
\label{sec:experiments:ours}
Our imputer uses the reconstructor plus test-time optimization pipeline from Sec.~\ref{sec:method}.
We instantiate $f_\theta$ as a standard reconstruction autoencoder, using different backbones (LSTM, MTAD-GAT~\cite{MTADGAT}, TCN-AE~\cite{thill2021temporal}, TransformerAE) that represent architectures from RNNs, CNNs, GNNs, and transformers. We define the baseline fill $B(X)$ as per-channel mean of the training set.
We tune our model hyperparameters with Optuna on the validation split using a dataset-agnostic search space, see Appendix~\ref{appendix-experiments-hyperparameter-optimization}.

\subsection{Test-time optimization details.}
\label{sec:experiments:tti_details}
Test-time input optimization runs for at most $K=5000$ iterations with per-sample early stopping.
We use Adam with learning rate $10^{-2}$.
Early stopping uses a bootstrap window of 100 iterations.
Letting $\bar{\mathcal{L}}_{\mathrm{init}}$ be the mean objective
over this window, we set an improvement threshold $\Delta_{\min}=0.01\,\bar{\mathcal{L}}_{\mathrm{init}}$ and a plateau
threshold $\tau=10^{-4}\,\bar{\mathcal{L}}_{\mathrm{init}}$.
A sample is marked converged if it fails to improve by $\Delta_{\min}$ for 100 consecutive iterations or if the
per-iteration objective change stays below $\tau$ for 200 consecutive iterations.
The MAE values in Table~\ref{tab:full_missing_test} are computed from the output selected by this objective-based stopping rule, not from an oracle choice of the iteration with the best test MAE.
We report three variants for our method: a pure forward pass without refinement (no optimization),
test-time optimization with $\lambda{=}0$, and test-time optimization with a simple first-order temporal smoothness regularizer $\mathcal{P}(V)$ on missing indices ($\lambda{=}10^{-3}$).
Reconstructor hyperparameters are chosen on validation data; the forward, no-prior, and prior variants are then all reported on the test set. The per-dataset \emph{best} rows used for the family comparison are post-hoc descriptive lower envelopes and never select the prediction for an individual test window.
Implementation and stopping details are summarized in Appendix~\ref{appendix-Algorithm}.

\subsection{Compute and reproducibility.}
\label{sec:experiments:compute}
Experiments were run in a containerized environment limited to 12 vCPUs and 48~GiB RAM with access to an
NVIDIA A30 MIG instance (16~GiB). We report mean $\pm$ standard deviation across 6 random seeds.

\begin{table*}[t]
\centering
\small
\setlength{\tabcolsep}{2.3pt}
\begin{tabular}{ll|ccccc}
\toprule
 & Setting & BeijingAir & ETTh1 & Electricity & ItalyAir & PeMS \\
\midrule
\multicolumn{7}{l}{\textit{Panel A: Conversion ablation for frozen reconstructors}} \\
\multirow{3}{*}{LSTM}
 & forward & 0.4968 $\pm$ 0.0145 & 0.6488 $\pm$ 0.0024 & 1.1901 $\pm$ 0.0389 & 0.5845 $\pm$ 0.0586 & 0.2944 $\pm$ 0.0064 \\
 & TTO, $\lambda{=}0$ & 0.3413 $\pm$ 0.0176 & 0.3286 $\pm$ 0.0216 & 1.1821 $\pm$ 0.0402 & 0.3806 $\pm$ 0.0365 & 0.2913 $\pm$ 0.0060 \\
 & TTO + prior & 0.3099 $\pm$ 0.0169 & 0.2898 $\pm$ 0.0416 & 1.1821 $\pm$ 0.0402 & 0.3796 $\pm$ 0.0365 & 0.2917 $\pm$ 0.0055 \\
\cline{1-7}
\multirow{3}{*}{MTAD-GAT}
 & forward & 0.2776 $\pm$ 0.0054 & 0.6227 $\pm$ 0.0203 & 1.5608 $\pm$ 0.0982 & 0.4942 $\pm$ 0.0128 & 0.3393 $\pm$ 0.0064 \\
 & TTO, $\lambda{=}0$ & 0.2726 $\pm$ 0.0042 & 0.5320 $\pm$ 0.0321 & 1.5580 $\pm$ 0.0990 & 0.4815 $\pm$ 0.0357 & 0.3378 $\pm$ 0.0062 \\
 & TTO + prior & 0.2728 $\pm$ 0.0051 & 0.5323 $\pm$ 0.0311 & 1.5580 $\pm$ 0.0990 & 0.4822 $\pm$ 0.0336 & 0.3377 $\pm$ 0.0061 \\
\cline{1-7}
\multirow{3}{*}{TCN-AE}
 & forward & 0.2455 $\pm$ 0.0039 & 0.6501 $\pm$ 0.0676 & 0.9853 $\pm$ 0.0287 & 0.7200 $\pm$ 0.0241 & 0.2898 $\pm$ 0.0030 \\
 & TTO, $\lambda{=}0$ & 0.2289 $\pm$ 0.0059 & 0.4299 $\pm$ 0.0347 & 0.9928 $\pm$ 0.0230 & 0.6557 $\pm$ 0.0268 & 0.2892 $\pm$ 0.0026 \\
 & TTO + prior & 0.2203 $\pm$ 0.0067 & 0.4329 $\pm$ 0.0307 & 0.9926 $\pm$ 0.0229 & 0.6404 $\pm$ 0.0241 & 0.2884 $\pm$ 0.0028 \\
\cline{1-7}
\multirow{3}{*}{TransformerAE}
 & forward & 0.2763 $\pm$ 0.0041 & 0.7610 $\pm$ 0.1168 & 1.0525 $\pm$ 0.0167 & 0.6440 $\pm$ 0.0399 & 0.2822 $\pm$ 0.0033 \\
 & TTO, $\lambda{=}0$ & 0.2255 $\pm$ 0.0043 & 0.6197 $\pm$ 0.1172 & 1.0354 $\pm$ 0.0220 & 0.4350 $\pm$ 0.0615 & 0.2811 $\pm$ 0.0051 \\
 & TTO + prior & 0.2200 $\pm$ 0.0034 & 0.5543 $\pm$ 0.1091 & 1.0353 $\pm$ 0.0221 & 0.4353 $\pm$ 0.0619 & 0.2816 $\pm$ 0.0055 \\
\midrule
\multicolumn{7}{l}{\textit{Panel B: Descriptive lower envelopes for external comparison}} \\
MeanFill & train mean & 0.7718 & 0.8043 & 1.9657 & 0.7974 & 0.7742 \\
Best matched mask-aware & per-dataset envelope & \textbf{0.1383 $\pm$ 0.0028} & 0.3222 $\pm$ 0.0261 & 1.3480 $\pm$ 0.0879 & \underline{0.3588 $\pm$ 0.0244} & \underline{0.2508 $\pm$ 0.0037} \\
Best OOD mask-aware & per-dataset envelope & \underline{0.2108 $\pm$ 0.0018} & \underline{0.3180 $\pm$ 0.0111} & \underline{1.1452 $\pm$ 0.0348} & \textbf{0.3488 $\pm$ 0.0087} & \textbf{0.2470 $\pm$ 0.0016} \\
Best refined reconstructor & per-dataset envelope & 0.2200 $\pm$ 0.0034 & \textbf{0.2898 $\pm$ 0.0416} & \textbf{0.9926 $\pm$ 0.0229} & 0.3796 $\pm$ 0.0365 & 0.2811 $\pm$ 0.0051 \\
\bottomrule
\end{tabular}
\caption{\textbf{Full-channel-missing evaluation (MAE; mean $\pm$ std where applicable).}
Panel A asks whether post-training conversion improves each frozen reconstructor. Panel B compares post-hoc per-dataset lower envelopes for matched-mask baselines, OOD structured-mask checkpoints, and refined reconstructors. Bold and underline mark the best and second-best Panel-B values. Every individual baseline checkpoint is reported in Appendix~\ref{appendix-experiments-full-channel-detailed}.}
\label{tab:full_missing_test}
\end{table*}

\subsection{Results}
\label{sec:experiments:results}

\paragraph{EQ1: Does conversion improve forward-only completion?}
Panel A of Table~\ref{tab:full_missing_test} makes the conversion ablation the primary result by comparing each frozen reconstructor with the same model after TTO. No-prior TTO ($\lambda=0$) lowers MAE in 19 of 20 backbone--dataset combinations. The sole exception is TCN-AE on Electricity, where MAE shows a $0.8\%$ relative increase. Using the reported means, averaging the pairwise relative MAE reduction
across the four backbones gives $14.6\%$ on BeijingAir, $29.1\%$ on ETTh1, $0.4\%$ on Electricity, $19.7\%$ on ItalyAir, and $0.5\%$ on PeMS, with a $12.9\%$ macro-average over all 20 pairs.

The smoothness prior is a separate, selective effect: relative to no-prior TTO, it lowers reported MAE in 11 of 20 combinations, leaves 2 unchanged at the shown precision, and raises MAE in 7. Thus, the 19/20 conversion result does not depend on uniform gains from the optional prior.

\paragraph{EQ2: How does conversion compare with mask-aware imputers?}
Panel B reports the lower mean MAE within each method family and dataset. Each cell in a \emph{best} row is the post-hoc minimum over the fully reported rows in that family: five matched full-channel-mask baselines, fifteen OOD point/subsequence/block checkpoints, or the eight refined reconstructor variants in Panel A, all individual baseline rows are traceable in Appendix~\ref{appendix-experiments-full-channel-detailed}.
Taking the refined-reconstructor envelope, TTO yields lower MAE than both mask-aware envelopes on ETTh1 and Electricity, but higher MAE than the strongest mask-aware result on BeijingAir, ItalyAir, and PeMS. Electricity requires a precise distinction: forward TCN-AE is the best reconstructor row, while the best \emph{refined} row is TCN-AE with the prior. That refined result still beats the OOD and matched-mask envelopes. Matched full-channel training substantially improves some baselines, but not uniformly. This comparison should retain the supervision boundary from Sec.~\ref{sec:experiments:missingness}. The mask-aware models receive artificial corruption during training, while the reconstructors are trained only on native missingness.

\paragraph{Mechanism diagnostics.}
The cross-Jacobian analysis in Sec.~\ref{sec:method:tti} establishes the local coupling condition behind observed-self-consistent refinement. Appendix~\ref{appendix-experiments-jacobian-diagnostics} reports illustrative diagnostics for three backbones across all five datasets, including stronger and weakly coupled cases.

\paragraph{EQ3: What optimization behavior and inference-time cost does conversion exhibit?}
Figure~\ref{fig:TTOconvergence} shows representative ETTh1 trajectories for all four reconstruction backbones. In this case, imputation MAE evolves smoothly rather than oscillating strongly, and the effect of the smoothness prior is most visible for LSTM and TransformerAE.

The hard cap $K=5000$ is therefore not the typical executed budget. For the fastest TTO rows used in the runtime comparison, median stopping is 201-206 iterations and the 90th percentile is 201-209. Although early stopping limits the realized iteration count, roughly 200 forward/backward updates per window still impose substantial computational overhead.

Using the per-dataset lower-envelope comparison defined in App.~\ref{app:clock_runtime}, TTO is \(6.7\)–-\(174.7\times\) slower than one-shot inference on identical hardware. This comparison uses the fastest measured configuration from each family and provides an estimate of TTO’s computational overhead. Complete per-dataset results are reported in Table~\ref{tab:runtime_normalized_tto_iterations}, while Figs.~\ref{fig:convergence-Beijing}-\ref{fig:convergence-PeMS} report convergence, active-sample counts, and runtime.
\section{Conclusion and outlook}
\label{sec:conclusion}

A frozen differentiable sequence reconstructor can often be converted into a point-estimate imputer without retraining its weights. Its learned temporal and cross-channel dependencies must make observed self-consistency informative about the missing inputs. TTO performs this conversion by optimizing only the missing entries of a test window through the frozen model while keeping observations clamped. We evaluate this claim in a controlled setting with five real-world datasets, four reconstruction-model families trained using only native missingness, and one complete channel synthetically masked in each test window. The experiments therefore demonstrate post-training conversion of these reconstructors.

Empirically, no-prior TTO lowers MAE relative to forward-only reconstruction in 19 of 20 backbone-dataset combinations, with a $12.9\%$ macro-average relative reduction (Table~\ref{tab:full_missing_test}). The exception is TCN-AE on Electricity, whose forward result is better than either refined variant. The optional smoothness prior is selectively beneficial as it improves 11 of 20 rows, is unchanged in 2 at reported precision, and worsens 7. Taking the best TTO-refined reconstructor per dataset, TTO yields lower MAE than the strongest mask-aware baseline on ETTh1 and Electricity, but higher MAE on BeijingAir, ItalyAir, and PeMS. This conditional accuracy benefit comes at cost of computation as stated in Sec.~\ref{sec:experiments:tti_details} and App.~\ref{app:clock_runtime}.

\paragraph{Limitations.}
The conversion fails at its mechanism boundary. For an identity-like reconstructor whose observed outputs do not depend on missing inputs, the missing-to-observed cross-Jacobian is zero and the observed residual supplies no recovery gradient. More generally, weak, ill-conditioned, or rank-deficient coupling leaves directions ambiguous, so multiple completions can be similarly self-consistent. Priors can regularize but cannot remove this ambiguity. Our Jacobian diagnostics provide mechanism-consistent evidence for three of the four backbones. TTO also introduces iterative compute and implementation choices such as optimizer, normalization, and stopping criteria, and the evaluation does not establish performance for arbitrary reconstructors.

\paragraph{Outlook.}
Future work should evaluate broader hard-missingness regimes, including multiple missing channels, mixed block+channel dropouts, and longer horizons. Stronger domain-aware priors, uncertainty diagnostics based on residual curvature or cross-Jacobian spectra, and faster quasi-Newton or Gauss-Newton-style solvers could make TTO more predictable and scalable. Given its dependence on reconstructor coupling and its iterative cost, TTO is best understood as a conditional inference-time adapter that complements, rather than replaces, amortized mask-aware imputers.
\paragraph{Impact}
This work studies test-time conversion of compatible frozen reconstructors into point-estimate imputers. Although reconstruction objectives also appear in anomaly detection, we do not evaluate deployed-model reuse, anomaly-detection performance, or maintenance savings. For sensor dropouts, imputed values used in downstream monitoring or forecasting should be flagged rather than treated as ground truth.

\bibliography{aaai2027}

\clearpage
\onecolumn
\appendix
\section{Cross-Jacobian coupling and local well-posedness}
\label{app:cross_jac}

This appendix supports the discussion in Sec.~\ref{sec:method:tti} by (i) making the dependence on the missing
coordinates explicit, (ii) deriving the cross-Jacobian identity in~\eqref{eq:method:cross_jac_def}, and (iii) stating
a simple local well-posedness implication.

\subsection{Vectorization and index operators}
\label{app:cross_jac:operators}
Let $M:=M^{\mathrm{fit}}$ be the fit-residual mask from Sec.~\ref{sec:problem_setup}, let $\Omega:=\{(t,d)\mid M_{t,d}=1\}$ be the fit-observed index set, and let $\bar\Omega$ denote the optimized index set.
Let $\mathrm{vec}:\mathbb{R}^{T\times D}\to\mathbb{R}^{TD}$ denote any fixed vectorization convention.

We use the standard linear maps induced by $\Omega$:
\begin{itemize}
\item A \emph{selection} matrix $S_\Omega\in\{0,1\}^{|\Omega|\times TD}$ such that $S_\Omega\,\mathrm{vec}(A)$ stacks the
entries of $A$ indexed by $\Omega$.
\item An \emph{embedding} matrix $E_{\bar\Omega}\in\{0,1\}^{TD\times|\bar\Omega|}$ such that $E_{\bar\Omega}v$ places
$v\in\mathbb{R}^{|\bar\Omega|}$ in the coordinates indexed by $\bar\Omega$ (and zeros elsewhere).
\end{itemize}
By construction, $S_\Omega E_{\bar\Omega}=0$.

Define the missing-coordinate optimization vector
\begin{equation}
v \;:=\; V_{\bar\Omega}\in\mathbb{R}^{|\bar\Omega|}.
\end{equation}
Then clamping observed entries and inserting missing values (Eq.~\eqref{eq:method:defs_V}) can be written as
\begin{equation}
\mathrm{vec}(X_V)
\;=\;
\mathrm{vec}(M\odot X)\;+\;E_{\bar\Omega}v.
\label{eq:app:Xv_linear_in_v}
\end{equation}
Eq.~\eqref{eq:app:Xv_linear_in_v} formalizes the statement in Sec.~\ref{sec:method:tti} that $X_V$ agrees with $X$ on
$\Omega$ while the coordinates in $\bar\Omega$ are free.

\subsection{Observed residual and cross-Jacobian identity}
\label{app:cross_jac:derivative}
Recall the masked residual and its observed stacking from Sec.~\ref{sec:method:tti}:
\begin{equation}
R(V)\;:=\;M\odot(f_\theta(X_V)-X),
\qquad
r(V)\;:=\;S_\Omega\,\mathrm{vec}(R(V))\in\mathbb{R}^{|\Omega|}.
\end{equation}
Since $S_\Omega$ selects only observed entries, the elementwise mask is redundant after selection:
\begin{equation}
r(V)
=
S_\Omega\,\mathrm{vec}\!\bigl(f_\theta(X_V)-X\bigr)
=
S_\Omega\,\mathrm{vec}\!\bigl(f_\theta(X_V)\bigr)\;-\;S_\Omega\,\mathrm{vec}(X).
\label{eq:app:r_simplify}
\end{equation}

Let $J_\theta(X_V)$ denote the Jacobian of the reconstructor output w.r.t.\ the vectorized input at $X_V$:
\begin{equation}
J_\theta(X_V)
\;:=\;
\frac{\partial\,\mathrm{vec}(f_\theta(U))}{\partial\,\mathrm{vec}(U)}\Big|_{U=X_V}
\in\mathbb{R}^{TD\times TD}.
\end{equation}
Applying the chain rule to \eqref{eq:app:r_simplify} and the affine map \eqref{eq:app:Xv_linear_in_v} yields
\begin{equation}
C(V)
\;:=\;
\frac{\partial r(V)}{\partial v}
\;=\;
S_\Omega\,J_\theta(X_V)\,E_{\bar\Omega}
\ \in\ \mathbb{R}^{|\Omega|\times|\bar\Omega|}.
\label{eq:app:cross_jac_formula}
\end{equation}
Eq.~\eqref{eq:app:cross_jac_formula} is the operator form of the cross-Jacobian block in~\eqref{eq:method:cross_jac_def}.

\subsection{Local isolation of self-consistent completions}
\label{app:cross_jac:wellposed}
Consider the noise-free self-consistency equation on observed entries,
\begin{equation}
r(V)=0,
\label{eq:app:self_consistency}
\end{equation}
and let $v^\star:=V^\star_{\bar\Omega}$ be a solution (i.e., $r(V^\star)=0$). Assume $f_\theta$ is differentiable in a
neighborhood of $X_{V^\star}$.

\begin{proposition}[Local isolation under full-column rank]
\label{prop:local_isolation}
If $C(V^\star)$ in \eqref{eq:app:cross_jac_formula} has full column rank, then $v^\star$ is an \emph{isolated} solution of
\eqref{eq:app:self_consistency}: there exists a neighborhood $\mathcal{U}$ of $v^\star$ such that
$r(V(v))=0$ and $v\in\mathcal{U}$ imply $v=v^\star$.
\end{proposition}

\begin{proof}[Proof sketch]
Work in the reduced variable $v$ via \eqref{eq:app:Xv_linear_in_v} and define $\phi(v):=r(V(v))$.
Full column rank of $C(V^\star)=\partial\phi/\partial v|_{v^\star}$ means $\phi$ is an immersion at $v^\star$.
By the constant-rank theorem, $\phi$ is locally an embedding around $v^\star$, hence locally injective.
Therefore $\phi(v)=\phi(v^\star)=0$ implies $v=v^\star$ in a sufficiently small neighborhood.
\end{proof}

\paragraph{Conditioning interpretation.}
For the least-squares data term $\tfrac12\|r(V)\|_2^2$, the Gauss--Newton approximation to the Hessian in $v$ is
$C(V)^\top C(V)$. Thus, if $C(V^\star)$ is ill-conditioned or rank-deficient, the self-consistency signal is weak or
ambiguous in some missing directions, motivating regularization through $\lambda\,\mathcal{P}(V)$ as described in
Sec.~\ref{sec:method:tti}.

\subsection{Minimal finite-dimensional example}
\label{app:cross_jac:example}
We give a concrete linear example illustrating $C(V)$ and why full column rank corresponds to informative coupling.

\paragraph{Setup.}
Let $T=2$ and $D=2$. Vectorize $x=\mathrm{vec}(X)\in\mathbb{R}^4$ as
$x=[x_1^{(1)},x_2^{(1)},x_1^{(2)},x_2^{(2)}]^\top$.
Assume channel~1 is observed at both times and channel~2 is missing at both times:
$\Omega=\{(1,1),(2,1)\}$ and $\bar\Omega=\{(1,2),(2,2)\}$, so $|\Omega|=|\bar\Omega|=2$.
Let $v=[v_1,v_2]^\top$ denote the missing coordinates.

\paragraph{A time-/channel-coupled linear reconstructor.}
Define $y=f_\theta(x)\in\mathbb{R}^4$ by
\begin{align}
y_1^{(1)} &= x_1^{(1)} + \alpha\,x_2^{(1)}, &
y_2^{(1)} &= x_2^{(1)}, \\
y_1^{(2)} &= x_1^{(2)} + \beta\,x_2^{(1)} + \gamma\,x_2^{(2)}, &
y_2^{(2)} &= x_2^{(2)}.
\end{align}
This encodes that the (missing) channel~2 influences the model’s prediction on the (observed) channel~1, including across
time via $\beta$.

\paragraph{Observed residual and cross-Jacobian.}
Clamping observed entries and inserting $v$ for the missing channel yields
\begin{equation}
r(V)=
\begin{bmatrix}
y_1^{(1)}-x_1^{(1)}\\
y_1^{(2)}-x_1^{(2)}
\end{bmatrix}
=
\begin{bmatrix}
\alpha\,v_1\\
\beta\,v_1+\gamma\,v_2
\end{bmatrix},
\qquad
C(V)=\frac{\partial r(V)}{\partial v}
=
\begin{bmatrix}
\alpha & 0\\
\beta & \gamma
\end{bmatrix}.
\end{equation}
Full column rank holds iff $\alpha\gamma\neq 0$, i.e., iff both missing coordinates locally affect the observed residual.
If $\alpha=0$, then $v_1$ is unidentifiable from observed self-consistency; if $\gamma=0$, then $v_2$ is unidentifiable.
This mirrors the intuition in Sec.~\ref{sec:method:tti}: test-time refinement is effective only when the reconstructor
induces sufficient coupling from missing to observed coordinates.

\clearpage
\section{Algorithm}
\label{appendix-Algorithm}
In our experiments we instantiate the optional prior $\mathcal{P}(V)$ as a first-order temporal smoothness penalty on the optimized entries.
We use the mask contract from Sec.~\ref{sec:problem_setup}: $M^{\mathrm{fit}}:=M^{\mathrm{nat}}\odot(1-M^{\mathrm{eval}})$ is the fit-residual mask and $O:=1-M^{\mathrm{fit}}$ is the optimization mask. The data term uses only $M^{\mathrm{fit}}$; baseline-filled native missing entries enter $O$, never the observations, while evaluator-only $H$ and $\tilde X$ never enter optimization.
Define the adjacent-validity mask
$W_\Delta\in\{0,1\}^{(T-1)\times D}$ by $(W_\Delta)_{t,d}=O_{t,d}\,O_{t-1,d}$, which activates only when both endpoints of a temporal difference are optimized.
We then use
\begin{equation}
\mathcal{P}(V)
:= \frac{1}{2(\|W_\Delta\|_1+\varepsilon)}
\bigl\|W_\Delta \odot (V_{2:T,:}-V_{1:T-1,:})\bigr\|_F^2,
\end{equation}
with $\varepsilon=10^{-8}$ for numerical stability.
In the batched implementation, $W_\Delta$ is computed per sample.
Full implementation details are summarized in Algorithm \ref{alg:tti}.
In our current implementation we keep the model in training mode, but disable dropout to be able to use cuDNN backward. This is the same as setting the model to evaluation mode, if no BatchNorm is used. If the model would use BatchNorm, set model to evaluation mode and set torch.backends.cudnn.flags(enabled=False). This is typically slower.

\begin{algorithm}[!t]
\caption{Test-time optimization of missing values (frozen reconstructor), batched (implementation-faithful)}
\label{alg:tti}
\begin{algorithmic}[1]
\Require baseline-filled batch $Y\in\mathbb{R}^{B\times T\times D}$ (observed entries intact), optimize-mask $O\in\{0,1\}^{B\times T\times D}$ (1 = optimize), fit-residual mask $M^{\mathrm{fit}}\in\{0,1\}^{B\times T\times D}$,
frozen reconstructor $f_\theta$, max iterations $K$, optimizer step size $\eta$, prior weight $\lambda$,
bootstrap window $W_0$, patience $p$, max-consecutive $q$, $\varepsilon>0$,
(optional) projection scale $\kappa$ with feature mean $\mu\in\mathbb{R}^D$ and std $\sigma\in\mathbb{R}^D$
\State $V \gets Y$ \Comment{optimization variable; same shape as $Y$}
\State $\mathrm{best}\gets (+\infty)^B,\ \mathrm{noImp}\gets \mathbf{0}^B,\ \mathrm{cons}\gets \mathbf{0}^B,\ \mathrm{conv}\gets \mathbf{0}\in\{0,1\}^B,\ \ell_{\mathrm{old}}\gets (+\infty)^B$
\State $\mathcal{L}_{\mathrm{init}}\gets [\ ]$,\quad $\Delta_{\min}\gets 0$,\quad $\tau\gets 0$
\State Initialize optimizer $\textsc{Opt}$ on $V$ (e.g., Adam with lr $\eta$)
\For{$k=0,\dots,K-1$}
  \State $\mathrm{act}\gets \mathbf{1}-\mathrm{conv}$ \Comment{active samples}
  \State $V[\neg O] \gets Y[\neg O]$ \Comment{clamp all fixed (non-optimized) entries}
  \State $\bar Y \gets f_\theta(V)$
  \State $n_F \gets \|M^{\mathrm{fit}}\|_1$ \Comment{per-sample count}
  \State $E_{\mathrm{fit}} \gets \tfrac{1}{2\,(n_F+\varepsilon)}\ \|M^{\mathrm{fit}}\odot(\bar Y-Y)\|_F^2$
  \State $W_\Delta \gets O_{:,2:T,:}\odot O_{:,1:T-1,:}$ \Comment{valid diffs: both endpoints optimized}
  \State $n_\Delta \gets \|W_\Delta\|_1$
  \State $\mathcal{P} \gets \tfrac{1}{2\,(n_\Delta+\varepsilon)}\ \|W_\Delta\odot(V[:,2:T,:]-V[:,1:T-1,:])\|_F^2$ \Comment{per-sample Frobenius norm}
  \State $\mathcal{J} \gets E_{\mathrm{fit}} + \lambda\,\mathcal{P}$ \Comment{per-sample objective}
  \State $\mathcal{J}_{\mathrm{tot}} \gets \sum_{b=1}^B \mathrm{act}_b\,\mathcal{J}_b$
  \State $\textsc{Opt}.\mathrm{zero\_grad}();\ \nabla_V \mathcal{J}_{\mathrm{tot}} \ \text{via backprop}$
  \State $G \gets (\nabla_V \mathcal{J}_{\mathrm{tot}}) \odot O$ \Comment{only update optimized entries}
  \State $G \gets G \odot \mathrm{act}$ \Comment{mask converged samples}
  \State $G \gets G / (\|G\|_2 + \varepsilon)$ \Comment{per-sample gradient normalization}
  \State Set $\nabla V \gets G$ and perform $\textsc{Opt}.\mathrm{step}()$
  \State $V[\neg O] \gets Y[\neg O]$ \Comment{re-clamp fixed entries after the step}
  \If{$\kappa$ is set}
    \State $\mathrm{lower}\gets \mu-\kappa\sigma,\ \mathrm{upper}\gets \mu+\kappa\sigma$
    \State $V[O] \gets \mathrm{clip}(V[O],\mathrm{lower},\mathrm{upper})$ \Comment{projection on optimized entries only}
  \EndIf
  \State $\delta \gets |\ell_{\mathrm{old}}-\mathcal{J}|,\quad \ell_{\mathrm{old}}\gets \mathcal{J}$
  \If{$k < W_0$}
    \State Append $\mathrm{mean}(\mathcal{J})$ to $\mathcal{L}_{\mathrm{init}}$
  \ElsIf{$k = W_0$}
    \State $\bar{\mathcal{J}}_{\mathrm{init}} \gets \mathrm{mean}(\mathcal{L}_{\mathrm{init}})$
    \State $\Delta_{\min}\gets 0.01\,\bar{\mathcal{J}}_{\mathrm{init}},\quad \tau\gets 0.01\,\Delta_{\min}$ \Comment{adaptive thresholds}
    \State $\mathrm{best}\gets \mathcal{J}$
  \Else
    \State $\mathrm{impr}\gets \mathcal{J} < (\mathrm{best}-\Delta_{\min})$
    \State $\mathrm{best}[\mathrm{impr}] \gets \mathcal{J}[\mathrm{impr}]$
    \State $\mathrm{noImp}[\neg \mathrm{impr}] \gets \mathrm{noImp}[\neg \mathrm{impr}] + 1$
    \State $\mathrm{good}\gets \delta > \tau$
    \State $\mathrm{cons}[\neg \mathrm{good}] \gets \mathrm{cons}[\neg \mathrm{good}] + 1;\ \mathrm{cons}[\mathrm{good}] \gets 0$
    \State $\mathrm{conv} \gets \mathrm{conv}\ \lor\ (\mathrm{noImp}\ge p)\ \lor\ (\mathrm{cons}\ge q)$
  \EndIf
\EndFor
\State $V[\neg O] \gets Y[\neg O]$,\quad $\bar Y \gets f_\theta(V)$
\State $\hat X \gets Y$;\quad $\hat X[O] \gets \bar Y[O]$ \Comment{replace optimized entries by model prediction}
\State \Return $\hat X$
\end{algorithmic}
\end{algorithm}

\clearpage
\section{Experiments}
\label{appendix-experiments}

\subsection{Benchmark models}
\label{appendix-experiments-benchmark-models}
We briefly summarize representative deep time-series models used as benchmarks or points of comparison.
\textit{iTransformer} \citep{liu2024itransformerinvertedtransformerseffective} adapts the Transformer to multivariate time series by \emph{inverting} the tokenization, i.e., treating variables as tokens (instead of time steps) so that attention primarily models cross-variate dependencies while temporal information is embedded in the token representations.
\textit{SAITS} \citep{SAITS_Du_2023} is a Transformer-based imputation model that uses diagonally masked self-attention blocks and combines multiple attention-based reconstructions to explicitly target missing-value recovery in multivariate sequences.
\textit{BRITS} \citep{BRITS_cao2018britsbidirectionalrecurrentimputation} performs imputation with bidirectional recurrent networks, propagating information forward and backward in time and iteratively refining estimates through consistency between the two directions.
\textit{TimesNet} \citep{wu2023timesnettemporal2dvariationmodeling} models time series via a 2D-style representation that captures multi-periodic temporal structure, enabling a unified architecture that has been applied across forecasting and other sequence tasks, including imputation.
\textit{FreTS} \citep{yi2023frequencydomain} operates in the frequency domain by transforming the input sequence and applying MLP-based modeling on spectral components, leveraging frequency-domain structure for time-series learning.

\subsection{Hyperparameter Optimization}
\label{appendix-experiments-hyperparameter-optimization}

For the backbone models, we used Optuna \cite{akiba2019optuna} to sample hyperparameters from the discrete search space in Table~\ref{tab:hpo_space_autoencoders}. Unless stated otherwise, categorical variables were sampled uniformly over the listed choices, and continuous variables were sampled uniformly over the given interval. The learning rate was sampled log-uniformly.
\begin{table}[t]
\centering
\small
\begin{tabular}{llll}
\toprule
& Hyperparameter & Type / Range & Notes \\
\midrule
\multicolumn{4}{l}{\textbf{LSTM}} \\
& $\texttt{hidden\_size}$
& \makecell[l]{categorical $\in \{16,32,64,96,$\\
$128,172,256,384,512,768,1024\}$}
& LSTM hidden dimension \\
& $\texttt{latent\_size}$
& \makecell[l]{categorical $\in \{4,6,8,12,16,32,64,$\\
$96,128,172,256,384,512,768,1024\}$}
& bottleneck dimension \\
& $\texttt{n\_layers}$ & integer $\in \{1,2,3,4,5\}$ & number of LSTM layers \\
& $\texttt{dropout}$ & float $\in \{0.00,0.05,0.10\}$ & LSTM dropout \\
& $\texttt{batch\_size}$ & categorical $\in \{32,64,128,256\}$ &  \\
& $\texttt{lr}$ & log-uniform in $[5\cdot 10^{-4},\,5\cdot 10^{-3}]$ & learning rate \\
& $\texttt{proj\_dim}$ & categorical $\in \{16,32,64,128,256\}$ & projection-layer width \\
& $\texttt{input\_proj\_dim}$ & categorical $\in \{16,32,64,128,256\}$ & input projection width \\
& $\texttt{post\_mlp\_ratio}$ & categorical $\in \{2,4,8\}$ & MLP expansion ratio \\
& $\texttt{post\_mlp\_dropout}$ & float $\in \{0.00,0.05,0.10\}$ & MLP dropout \\
\midrule
\multicolumn{4}{l}{\textbf{TCNAE}} \\
& $\texttt{channel\_preset}$
& \makecell[l]{categorical $\in \{$tiny, small\_tight, small,\\
medium\_tight, medium\_plus, \\
large\_tight, large$\}$}
& sets base and latent channels \\
& $\texttt{latent\_pool}$ & categorical $\in \{2,3,4,6,8,12\}$ & temporal bottleneck pooling \\
& $\texttt{kernel\_size}$ & categorical $\in \{3,5,7,9\}$ & temporal convolution kernel \\
& $\texttt{dilation\_preset}$ & \makecell[l]{categorical $\in \{$local, default, deep,\\
 extra\_deep$\}$} & encoder/decoder dilation schedule \\
& $\texttt{dropout}$ & categorical $\in \{0.00,0.05,0.10,0.20\}$ & convolutional dropout \\
& $\texttt{batch\_size}$ & categorical $\in \{1,2,4,8,16,32,64\}$ & dataset-dependent subset \\
& $\texttt{lr}$ & log-uniform in $[10^{-4},\,5\cdot 10^{-3}]$ & dataset-dependent upper bound \\
\midrule
\multicolumn{4}{l}{\textbf{TransformerAE}} \\
& $\texttt{d\_model}$ & categorical $\in \{32,64,96,128\}$ & transformer width \\
& $\texttt{n\_heads}$ & categorical $\in \{2,4,8\}$ & attention heads \\
& $\texttt{latent\_dim\_ratio}$ & categorical $\in \{0.125,0.25,0.5\}$ & determines latent dimension \\
& $\texttt{ff\_ratio}$ & categorical $\in \{2,4\}$ & encoder/decoder FFN expansion \\
& $\texttt{latent\_ff\_ratio}$ & categorical $\in \{2,4\}$ & latent FFN expansion \\
& $\texttt{encoder\_depth}$ & categorical $\in \{1,2,3\}$ & encoder blocks \\
& $\texttt{latent\_depth}$ & categorical $\in \{1,2\}$ & latent transformer blocks \\
& $\texttt{decoder\_depth}$ & categorical $\in \{1,2,3\}$ & decoder blocks \\
& $\texttt{downsample\_factor}$ & categorical $\in \{2,3,4,6,8,12\}$ & temporal compression \\
& $\texttt{dropout}$ & categorical $\in \{0.00,0.05,0.10,0.20\}$ & transformer dropout \\
& $\texttt{stem\_kernel\_size}$ & categorical $\in \{1,3,5\}$ & input convolution kernel \\
& $\texttt{output\_head\_kernel\_size}$ & categorical $\in \{1,3\}$ & output head kernel \\
& $\texttt{upsample\_mode}$ & categorical $\in \{$linear, nearest$\}$ & decoder upsampling \\
& $\texttt{batch\_size}$ & categorical $\in \{1,2,4,8,16,32,64\}$ & dataset-dependent subset \\
& $\texttt{lr}$ & log-uniform in $[10^{-4},\,5\cdot 10^{-3}]$ & dataset-dependent upper bound \\
\midrule
\multicolumn{4}{l}{\textbf{MTAD-GAT}} \\
& $\texttt{kernel\_size}$ & categorical $\in \{1,3,5,7,9\}$ & temporal convolution kernel \\
& $\texttt{feat\_gat\_embed\_dim}$ & categorical $\in \{2,4,8,16,24\}$ & feature-attention embedding \\
& $\texttt{time\_gat\_embed\_dim}$ & categorical $\in \{4,8,16,24,32\}$ & temporal-attention embedding \\
& $\texttt{gru\_n\_layers}$ & categorical $\in \{1,2\}$ & GRU depth \\
& $\texttt{gru\_hid\_dim}$ & categorical $\in \{8,16,32,48,64,96,128\}$ & GRU hidden dimension \\
& $\texttt{latent\_dim\_ratio}$ & categorical $\in \{0.125,0.25,0.5\}$ & determines latent dimension \\
& $\texttt{recon\_n\_layers}$ & categorical $\in \{1,2\}$ & reconstruction MLP depth \\
& $\texttt{recon\_hid\_dim}$ & categorical $\in \{8,16,32,48,64,96,128\}$ & reconstruction hidden width \\
& $\texttt{dropout}$ & categorical $\in \{0.00,0.05,0.10,0.20\}$ & model dropout \\
& $\texttt{alpha}$ & categorical $\in \{0.05,0.10,0.20,0.30\}$ & LeakyReLU slope in GAT layers \\
& $\texttt{batch\_size}$ & categorical $\in \{1,2,4,8,16,32,64\}$ & dataset-dependent subset \\
& $\texttt{lr}$ & log-uniform in $[10^{-4},\,5\cdot 10^{-3}]$ & dataset-dependent upper bound \\
\midrule
\multicolumn{4}{l}{\textbf{Fixed training settings}} \\
& $\texttt{epochs}$ & fixed $=100$ & maximum training epochs \\
& $\texttt{patience}$ & fixed $=50$ & early-stopping patience \\
& $\texttt{n\_trials}$ & fixed $=20$ for backbone autoencoders & Optuna trials per model--dataset pair \\
\bottomrule
\end{tabular}
\caption{Optuna hyperparameter search spaces for the LSTM and backbone autoencoder models. For TCN-AE, TransformerAE, and MTAD-GAT, reported ranges are the union over datasets; the implementation uses dataset-specific subsets to account for sequence length and memory constraints.}
\label{tab:hpo_space_autoencoders}
\end{table}

\subsection{Native Missingness}
\label{appendix-experiments-native-missingness}
Table~\ref{tab:native_missingness} reports the dimensionality, window length, and native missingness rate for each dataset before any artificial corruption is applied.
This distinction matters because the frozen reconstructors are trained only on the native training data, while the benchmark evaluation masks are added synthetically at test time.
\begin{table}[t]
\centering
\small
\begin{tabular}{lccc}
\toprule
Dataset & $D$ & $T$ (window) & Native missing rate \\
\midrule
BeijingAir    & 132 & 24 & 1.60\% \\
ItalyAir      & 13  & 12 & 0\% \\
ETTh1        & 7   & 48 & 0\% \\
Electricity   & 370 & 96 & 0\% \\
PeMS          & 862 & 24 & 0\% \\
\bottomrule
\end{tabular}
\caption{\textbf{Dataset characteristics and native missingness.}
Native missingness rates refer to the raw series prior to any artificial corruption.
$D$ is the number of variables and $T$ is the non-overlapping window length used to generate samples, following the preprocessing summary in TSI-Bench~\cite{du2024tsibenchbenchmarkingtimeseries}.}
\label{tab:native_missingness}
\end{table}

\subsection{Detailed Full-Channel-Missing Results}
\label{appendix-experiments-full-channel-detailed}
Table~\ref{tab:full_missing_detailed} reports every matched-mask and OOD structured-mask checkpoint underlying the compact family-level comparison in the main paper. The individual rows are retained here so that the descriptive lower envelopes in Table~\ref{tab:full_missing_test} can be traced to their source methods and training-mask regimes.
\begin{table*}[t]
\centering
\small
\resizebox{\textwidth}{!}{\begin{tabular}{ll|ccccc}
\toprule
            & Setting & BeijingAir & ETTh1 & Electricity & ItalyAir & PeMS \\
\midrule
\multicolumn{7}{l}{\textit{Simple statistical baseline}} \\
MeanFill
            & train mean & 0.7718 & 0.8043 & 1.9657 & 0.7974 & 0.7742 \\
\midrule
\midrule
\multicolumn{7}{l}{\textit{Matched full-channel-mask baselines}} \\
BRITS
            & full feature & \textbf{0.1383 $\pm$ 0.0028} & 0.9848 $\pm$ 0.1342 & 1.3824 $\pm$ 0.0316 & 0.3588 $\pm$ 0.0244 & 0.2690 $\pm$ 0.0032 \\
FreTS
            & full feature & 0.1655 $\pm$ 0.0024 & 0.6003 $\pm$ 0.0359 & 1.3480 $\pm$ 0.0879 & 0.4360 $\pm$ 0.0318 & 0.3284 $\pm$ 0.0096 \\
SAITS
            & full feature & \underline{0.1575 $\pm$ 0.0053} & 0.3222 $\pm$ 0.0261 & 1.6103 $\pm$ 0.0193 & 0.3648 $\pm$ 0.0197 & 0.2508 $\pm$ 0.0037 \\
TimesNet
            & full feature & 0.7039 $\pm$ 0.0024 & 0.8059 $\pm$ 0.0021 & 1.9124 $\pm$ 0.0827 & 0.5522 $\pm$ 0.0094 & 0.6518 $\pm$ 0.0192 \\
iTransformer
            & full feature & 0.7634 $\pm$ 0.0048 & 0.8373 $\pm$ 0.0032 & 1.9117 $\pm$ 0.0480 & 0.7981 $\pm$ 0.0006 & 0.6733 $\pm$ 0.0325 \\
\midrule
\midrule
\multicolumn{7}{l}{\textit{OOD transfer from standard structured-mask training}} \\
\multirow{3}{*}{BRITS}
            & block  & 0.2108 $\pm$ 0.0018 & \underline{0.3180 $\pm$ 0.0111} & 1.2736 $\pm$ 0.0307 & 0.3665 $\pm$ 0.0133 & 0.2485 $\pm$ 0.0024 \\
            & point  & 0.2115 $\pm$ 0.0019 & 0.3227 $\pm$ 0.0129 & 1.2963 $\pm$ 0.0620 & 0.3740 $\pm$ 0.0144 & \underline{0.2472 $\pm$ 0.0033} \\
            & subseq & 0.2128 $\pm$ 0.0033 & 0.3287 $\pm$ 0.0124 & 1.2555 $\pm$ 0.0126 & 0.3797 $\pm$ 0.0184 & 0.2494 $\pm$ 0.0026 \\
\cline{1-7}
\multirow{3}{*}{FreTS}
            & block  & 0.2371 $\pm$ 0.0122 & 0.4330 $\pm$ 0.0395 & 1.1770 $\pm$ 0.0879 & 0.4054 $\pm$ 0.0397 & 0.3743 $\pm$ 0.0069 \\
            & point  & 0.2431 $\pm$ 0.0200 & 0.4322 $\pm$ 0.0288 & 1.1452 $\pm$ 0.0348 & 0.4137 $\pm$ 0.0338 & 0.3773 $\pm$ 0.0113 \\
            & subseq & 0.2418 $\pm$ 0.0185 & 0.4302 $\pm$ 0.0343 & 1.1581 $\pm$ 0.0754 & 0.4122 $\pm$ 0.0333 & 0.3732 $\pm$ 0.0076 \\
\cline{1-7}
\multirow{3}{*}{SAITS}
            & block  & 0.2233 $\pm$ 0.0020 & 0.3546 $\pm$ 0.0109 & 1.6379 $\pm$ 0.0675 & \textbf{0.3488 $\pm$ 0.0087} & 0.2501 $\pm$ 0.0046 \\
            & point  & 0.2276 $\pm$ 0.0066 & 0.3543 $\pm$ 0.0072 & 1.6135 $\pm$ 0.0465 & \underline{0.3513 $\pm$ 0.0105} & 0.2481 $\pm$ 0.0025 \\
            & subseq & 0.2243 $\pm$ 0.0031 & 0.3545 $\pm$ 0.0066 & 1.6170 $\pm$ 0.0732 & 0.3501 $\pm$ 0.0088 & \textbf{0.2470 $\pm$ 0.0016} \\
\cline{1-7}
\multirow{3}{*}{TimesNet}
            & block  & 0.4501 $\pm$ 0.0074 & 0.7612 $\pm$ 0.0108 & 1.7867 $\pm$ 0.0142 & 0.5805 $\pm$ 0.0048 & 0.2666 $\pm$ 0.0025 \\
            & point  & 0.4471 $\pm$ 0.0101 & 0.7624 $\pm$ 0.0103 & 1.8161 $\pm$ 0.0526 & 0.5798 $\pm$ 0.0054 & 0.2678 $\pm$ 0.0037 \\
            & subseq & 0.4582 $\pm$ 0.0124 & 0.7637 $\pm$ 0.0097 & 1.7747 $\pm$ 0.0101 & 0.5812 $\pm$ 0.0047 & 0.2663 $\pm$ 0.0019 \\
\cline{1-7}
\multirow{3}{*}{iTransformer}
            & block  & 0.7631 $\pm$ 0.0032 & 0.8072 $\pm$ 0.0052 & 1.9836 $\pm$ 0.0429 & 0.7980 $\pm$ 0.0014 & 0.5479 $\pm$ 0.0220 \\
            & point  & 0.7634 $\pm$ 0.0040 & 0.8062 $\pm$ 0.0033 & 2.0083 $\pm$ 0.1209 & 0.7980 $\pm$ 0.0013 & 0.5596 $\pm$ 0.0118 \\
            & subseq & 0.7613 $\pm$ 0.0030 & 0.8047 $\pm$ 0.0023 & 1.9637 $\pm$ 0.0382 & 0.7976 $\pm$ 0.0011 & 0.5810 $\pm$ 0.0342 \\
\midrule
\midrule
\multicolumn{7}{l}{\textit{Frozen reconstructors with test-time optimization}} \\
\multirow{3}{*}{LSTM}
            & forward (no opt)                       & 0.4968 $\pm$ 0.0145 & 0.6488 $\pm$ 0.0024 & 1.1901 $\pm$ 0.0389 & 0.5845 $\pm$ 0.0586 & 0.2944 $\pm$ 0.0064 \\
            & opt, $\lambda{=}0$ (no prior)          & 0.3413 $\pm$ 0.0176 & 0.3286 $\pm$ 0.0216 & 1.1821 $\pm$ 0.0402 & 0.3806 $\pm$ 0.0365 & 0.2913 $\pm$ 0.0060 \\
            & opt + smooth prior ($\lambda{=}10^{-3}$) & 0.3099 $\pm$ 0.0169 & \textbf{0.2898 $\pm$ 0.0416} & 1.1821 $\pm$ 0.0402 & 0.3796 $\pm$ 0.0365 & 0.2917 $\pm$ 0.0055 \\
\cline{1-7}
\multirow{3}{*}{MTAD-GAT}
            & forward (no opt)                       & 0.2776 $\pm$ 0.0054 & 0.6227 $\pm$ 0.0203 & 1.5608 $\pm$ 0.0982 & 0.4942 $\pm$ 0.0128 & 0.3393 $\pm$ 0.0064 \\
            & opt, $\lambda{=}0$ (no prior)          & 0.2726 $\pm$ 0.0042 & 0.5320 $\pm$ 0.0321 & 1.5580 $\pm$ 0.0990 & 0.4815 $\pm$ 0.0357 & 0.3378 $\pm$ 0.0062 \\
            & opt + smooth prior ($\lambda{=}10^{-3}$) & 0.2728 $\pm$ 0.0051 & 0.5323 $\pm$ 0.0311 & 1.5580 $\pm$ 0.0990 & 0.4822 $\pm$ 0.0336 & 0.3377 $\pm$ 0.0061 \\
\cline{1-7}
\multirow{3}{*}{TCN-AE}
            & forward (no opt)                       & 0.2455 $\pm$ 0.0039 & 0.6501 $\pm$ 0.0676 & \textbf{0.9853 $\pm$ 0.0287} & 0.7200 $\pm$ 0.0241 & 0.2898 $\pm$ 0.0030 \\
            & opt, $\lambda{=}0$ (no prior)          & 0.2289 $\pm$ 0.0059 & 0.4299 $\pm$ 0.0347 & 0.9928 $\pm$ 0.0230 & 0.6557 $\pm$ 0.0268 & 0.2892 $\pm$ 0.0026 \\
            & opt + smooth prior ($\lambda{=}10^{-3}$) & 0.2203 $\pm$ 0.0067 & 0.4329 $\pm$ 0.0307 & \underline{0.9926 $\pm$ 0.0229} & 0.6404 $\pm$ 0.0241 & 0.2884 $\pm$ 0.0028 \\
\cline{1-7}
\multirow{3}{*}{TransformerAE}
            & forward (no opt)                       & 0.2763 $\pm$ 0.0041 & 0.7610 $\pm$ 0.1168 & 1.0525 $\pm$ 0.0167 & 0.6440 $\pm$ 0.0399 & 0.2822 $\pm$ 0.0033 \\
            & opt, $\lambda{=}0$ (no prior)          & 0.2255 $\pm$ 0.0043 & 0.6197 $\pm$ 0.1172 & 1.0354 $\pm$ 0.0220 & 0.4350 $\pm$ 0.0615 & 0.2811 $\pm$ 0.0051 \\
            & opt + smooth prior ($\lambda{=}10^{-3}$) & 0.2200 $\pm$ 0.0034 & 0.5543 $\pm$ 0.1091 & 1.0353 $\pm$ 0.0221 & 0.4353 $\pm$ 0.0619 & 0.2816 $\pm$ 0.0055 \\
\bottomrule
\end{tabular}
}\caption{\textbf{Detailed full-channel-missing evaluation (MAE; mean $\pm$ std where applicable).}
One channel is fully masked at test time, in addition to native missingness, with balanced cyclic channel assignment.
MeanFill is a deterministic per-channel training-mean imputer.
\emph{Full feature} denotes matched channel-drop training; \emph{point}, \emph{subseq}, and \emph{block} denote OOD transfer from 10\% structured-mask training.
Frozen reconstructors are evaluated with forward-only inference and TTO. Best results are marked in bold; second-best results are underlined. This table expands the compact family-level comparison in the main paper and preserves every individual checkpoint row.}
\label{tab:full_missing_detailed}
\end{table*}

\subsection{Trainable Parameters}
\label{appendix-experiments-trainable-parameters}
Table~\ref{tab:trainable_params} compares trainable parameter counts for the mask-aware baselines and the reconstruction backbones used with test-time optimization.
We retain these counts as supplementary capacity context rather than as a core contribution or scaling claim; several TTO backbones are smaller than the largest deep imputation baselines, but model size is not the focus of the conversion study.
\begin{table}[t]
\centering
\small
\resizebox{\linewidth}{!}{\begin{tabular}{l|ccccc}
\toprule
 & BeijingAir & ETTh1 & Electricity & ItalyAir & PeMS \\
\midrule
BRITS        &   3,598,496      &   2,178,496       &  17,082,800       &     596,912       &  32,012,048 \\
FreTS        &     909,852      &     465,271       &   3,706,194       &     668,313       &   1,715,958 \\
SAITS        &   7,153,808      &  88,235,470       &  63,624,720       &  16,628,642       &  78,229,072 \\
TimesNet     &  87,063,940      &   5,510,663       &  45,569,394       &  22,051,853       &  91,622,238 \\
iTransformer &   8,286,232      &  23,723,056       &  12,989,024       &  18,932,236       &   1,854,744 \\
\hline
TCN-AE       &   1,030,052      &   1,000,999       &     513,810       &  \textit{39,965}  &  \textbf{461,822} \\
TransformerAE& \textbf{338,676} &   1,075,975       &     768,322       &      89,997       &  \textit{650,182} \\
MTAD-GAT     &  \textit{446,624}&  \textit{106,188} &     977,936       &  \textbf{32,156}  &   1,774,276 \\
LSTM         &     873,492      & \textbf{12,669}   & \textbf{42,930}   &     494,781       &     894,142 \\
\bottomrule
\end{tabular}
}\caption{Trainable parameters per model and dataset. Bold indicates the smallest model for each dataset; italics indicate the second-smallest model.}
\label{tab:trainable_params}
\end{table}

\subsection{Jacobian diagnostics for test-time optimization}
\label{appendix-experiments-jacobian-diagnostics}
The analysis in Appendix~\ref{app:cross_jac} identifies the cross-Jacobian
$C(V)=\partial r(V)/\partial v$ as the local object that determines whether changes in the optimized missing
coordinates are visible through the observed self-consistency residual.
To check this mechanism empirically, we ran Jacobian diagnostics on the full-channel-missing setting for
LSTM, TCN-AE, and TransformerAE across all five datasets, comparing TTO with $\lambda{=}0$ against TTO with the
smoothness prior $\lambda{=}10^{-3}$. MTAD-GAT~\cite{MTADGAT} was omitted because of its substantially higher diagnostic runtime.
For each model dataset variant pair, we used the selected checkpoint round from the corresponding TTO run and
evaluated 32 test windows, recording the mean MAE, the smallest singular value of $C(V)$, the spectral condition
number, and the Gauss--Newton curvature range along the optimization trajectory.

In this setting, the condition number is
\begin{equation}
\kappa(C(V))=\frac{\sigma_{\max}(C(V))}{\sigma_{\min}(C(V))},
\end{equation}
where $C(V)$ maps perturbations of the optimized missing entries to perturbations of the observed residual.
Small $\sigma_{\min}$ or large $\kappa(C(V))$ indicate directions in the missing channel that are only weakly
constrained by observed self-consistency; in such directions, the least-squares curvature
$C(V)^\top C(V)$ is flat or anisotropic, so optimization can be slow, sensitive, or biased toward arbitrary
completions.
Moderate condition numbers with non-negligible $\sigma_{\min}$ instead indicate that the reconstructor couples the
missing channel to the observed channels strongly enough for TTO to receive informative gradients.
The smoothness prior does not change $C(V)$ itself; rather, it adds an external regularizing term to the objective,
which can stabilize underconstrained directions when the data term alone is weak.

\begin{figure}[t]
    \centering
    \includegraphics[width=0.92\linewidth]{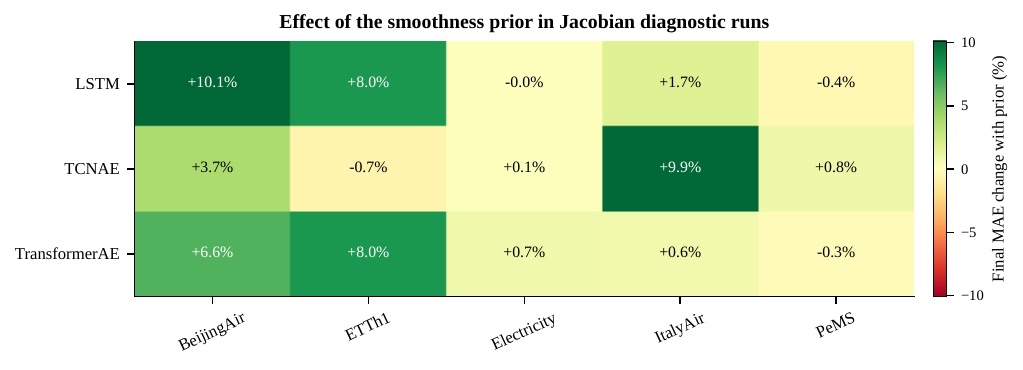}
    \caption{\textbf{Effect of the smoothness prior in the Jacobian diagnostic runs.}
    Each cell reports the percentage change in final diagnostic MAE when adding the smoothness prior
    ($\lambda{=}10^{-3}$) relative to TTO without the prior ($\lambda{=}0$), for the same model, dataset, and selected
    checkpoint round. Positive values indicate that the prior reduces MAE. The strongest gains occur in several
    BeijingAir and ETTh1 settings, while PeMS and TCN-AE/ETTh1 show small negative or near-neutral changes.}
    \label{fig:jacobian-prior-improvement-heatmap}
\end{figure}

\begin{figure}[t]
    \centering
    \includegraphics[width=0.95\linewidth]{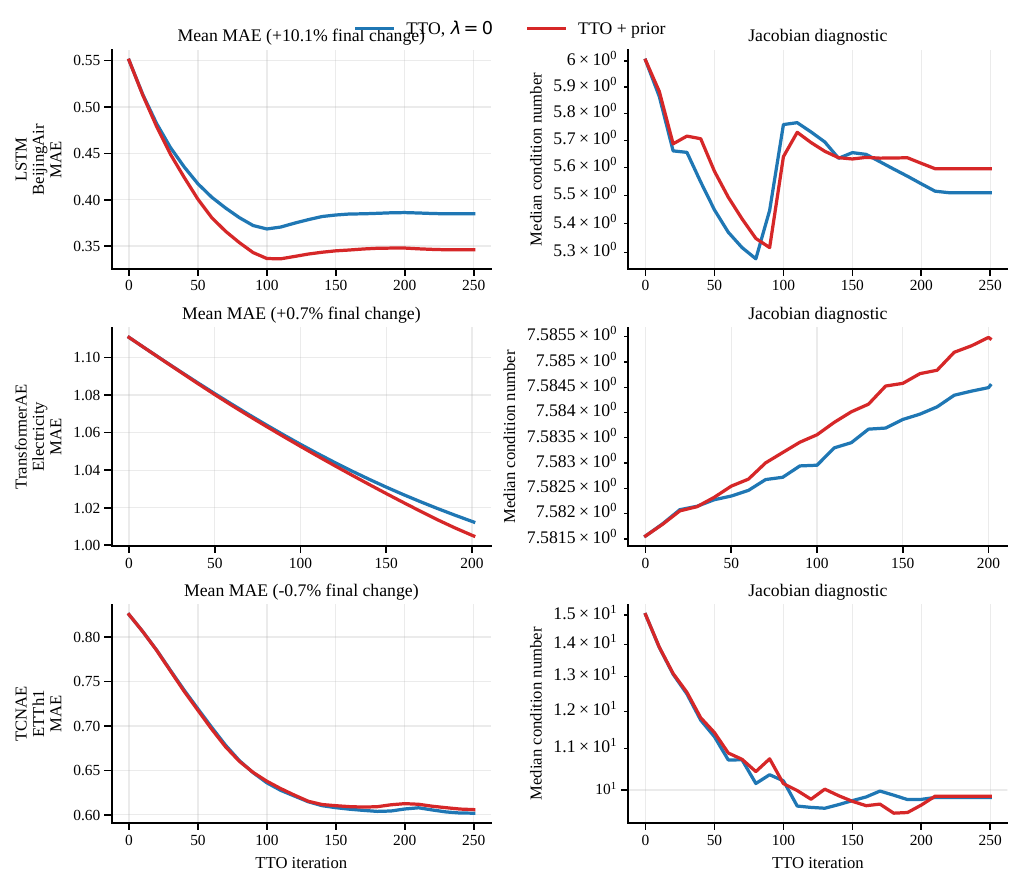}
    \caption{\textbf{Representative TTO trajectories with Jacobian diagnostics.}
    For three model--dataset pairs, the left column shows mean evaluation MAE over diagnostic TTO iterations and the
    right column shows the median cross-Jacobian condition number $\kappa(C(V))$ on a log scale. Blue curves correspond
    to TTO without the prior and red curves to TTO with the smoothness prior. These traces connect the accuracy changes
    in Table~\ref{tab:jacobian_diagnostics_summary} to the local conditioning of the observed self-consistency map:
    successful refinement reduces MAE while maintaining a usable coupling between optimized missing entries and
    observed residuals.}
    \label{fig:jacobian-representative-traces}
\end{figure}

Table~\ref{tab:jacobian_diagnostics_summary} summarizes the paired diagnostic runs.
The prior improves the final diagnostic MAE in 11 of 15 model--dataset pairs, with the largest gains on
LSTM/BeijingAir, LSTM/ETTh1, TCN-AE/ItalyAir, TransformerAE/BeijingAir, and TransformerAE/ETTh1.
The few negative cases are small, which is consistent with treating the prior as a stabilizer rather than as a
uniformly beneficial accuracy mechanism.
Figure~\ref{fig:jacobian-prior-improvement-heatmap} visualizes the same paired effect as a model-by-dataset heatmap.
Figure~\ref{fig:jacobian-representative-traces} shows representative optimization trajectories: the left panels track
the mean MAE over TTO iterations, while the right panels track the Jacobian condition number on a log scale.
The representative plots are consistent with the mechanism that useful missing-to-observed coupling permits informative refinement, while weak or anisotropic coupling can leave directions underconstrained. They are illustrative diagnostics and do not establish a systematic association by which condition number alone predicts imputation accuracy or prior benefit.

\begin{table}[t]
\centering
\small
\resizebox{\linewidth}{!}{\begin{tabular}{llrccrrrr}
\toprule
Model & Dataset & Round & MAE, $\lambda{=}0$ & MAE, prior & $\Delta$ final (\%) & $\Delta$ best (\%) & $\sigma_{\min}$, prior & Cond. ratio \\
\midrule
LSTM & BeijingAir & 5 & 0.3849 & 0.3461 & \textbf{+10.1} & \textbf{+8.7} & 0.0856 & 1.02 \\
LSTM & ETTh1 & 2 & 0.3523 & 0.3241 & \textbf{+8.0} & \textbf{+8.0} & 0.1943 & 1.00 \\
LSTM & Electricity & 1 & 1.0405 & 1.0405 & -0.0 & +0.0 & 0.0891 & 1.00 \\
LSTM & ItalyAir & 0 & 0.3701 & 0.3638 & \textbf{+1.7} & \textbf{+1.7} & 0.2317 & 1.02 \\
LSTM & PeMS & 1 & 0.2916 & 0.2929 & -0.4 & -0.4 & 6.23e-03 & 0.89 \\
\midrule
TCNAE & BeijingAir & 4 & 0.2992 & 0.2881 & \textbf{+3.7} & \textbf{+1.7} & 0.0145 & 1.04 \\
TCNAE & ETTh1 & 0 & 0.6019 & 0.6059 & -0.7 & -0.7 & 0.0605 & 1.00 \\
TCNAE & Electricity & 0 & 1.0434 & 1.0423 & \textbf{+0.1} & +0.0 & 1.03e-03 & 1.04 \\
TCNAE & ItalyAir & 0 & 0.6632 & 0.5973 & \textbf{+9.9} & \textbf{+5.3} & 0.0747 & 1.03 \\
TCNAE & PeMS & 3 & 0.2943 & 0.2919 & \textbf{+0.8} & \textbf{+0.8} & 1.13e-04 & 1.00 \\
\midrule
TransformerAE & BeijingAir & 3 & 0.2811 & 0.2627 & \textbf{+6.6} & \textbf{+6.6} & 0.1165 & 0.98 \\
TransformerAE & ETTh1 & 0 & 0.9656 & 0.8888 & \textbf{+8.0} & \textbf{+1.5} & 0.0394 & 0.95 \\
TransformerAE & Electricity & 5 & 1.0124 & 1.0049 & \textbf{+0.7} & \textbf{+0.7} & 0.0438 & 1.00 \\
TransformerAE & ItalyAir & 0 & 0.3979 & 0.3954 & \textbf{+0.6} & \textbf{+0.6} & 0.1409 & 0.98 \\
TransformerAE & PeMS & 4 & 0.3048 & 0.3057 & -0.3 & -0.3 & 0.0622 & 0.97 \\
\bottomrule
\end{tabular}
}
\caption{\textbf{Jacobian diagnostic summary for TTO with and without the smoothness prior.} Each row compares paired diagnostic runs for the selected checkpoint round. $\Delta$ final and $\Delta$ best report the percentage MAE reduction from adding the prior (positive values are improvements). The condition ratio is the final median Jacobian condition number with prior divided by the corresponding value without prior.}
\label{tab:jacobian_diagnostics_summary}
\end{table}

\subsection{Results on Benchmark Missingness}
\label{appendix-experiments-benchmark-missingness-results}
Tables~\ref{tab:normal_test_point}, \ref{tab:normal_test_subseq}, and \ref{tab:normal_test_block} report in-distribution MAE on the standard TSI-Bench point, subsequence, and block missingness settings.
They complement the main full-channel-missing table by showing how the same benchmark baselines and frozen reconstructors behave when the test masks match the usual structured missingness regimes.

\begin{table*}[t]
\centering
\small
\resizebox{\textwidth}{!}{\begin{tabular}{ll|ccccc}
\toprule
             & Setting & BeijingAir & ETTh1 & Electricity & ItalyAir & PeMS \\
\midrule
MeanFill
             & train mean & 0.7563 & 0.8305 & 1.8667 & 0.7716 & 0.8037 \\
\midrule
BRITS
             & point & \underline{0.1279 $\pm$ 0.0020} & \underline{0.1550 $\pm$ 0.0041} & 0.9704 $\pm$ 0.0182 & 0.2269 $\pm$ 0.0059 & \underline{0.2605 $\pm$ 0.0019} \\
FreTS
             & point & 0.2012 $\pm$ 0.0091 & 0.2069 $\pm$ 0.0071 & \textbf{0.6853 $\pm$ 0.1011} & 0.2362 $\pm$ 0.0158 & 0.3515 $\pm$ 0.0145 \\
SAITS
             & point & 0.1550 $\pm$ 0.0042 & \textbf{0.1215 $\pm$ 0.0040} & 1.4574 $\pm$ 0.0146 & 0.1707 $\pm$ 0.0083 & 0.2656 $\pm$ 0.0010 \\
TimesNet
             & point & 0.4098 $\pm$ 0.0077 & 0.7260 $\pm$ 0.0080 & 1.7587 $\pm$ 0.0131 & 0.5349 $\pm$ 0.0095 & 0.2814 $\pm$ 0.0025 \\
iTransformer
             & point & \textbf{0.1046 $\pm$ 0.0024} & 0.2623 $\pm$ 0.0066 & \underline{0.8429 $\pm$ 0.1469} & 0.1806 $\pm$ 0.0032 & \textbf{0.1490 $\pm$ 0.0110} \\
\midrule
\midrule
\multirow{3}{*}{LSTM}
             & forward (no opt)                       & 0.5111 $\pm$ 0.0116 & 0.6856 $\pm$ 0.0023 & 1.1178 $\pm$ 0.0355 & 0.5474 $\pm$ 0.0387 & 0.3198 $\pm$ 0.0037 \\
             & opt, $\lambda{=}0$ (no prior)          & 0.4317 $\pm$ 0.0063 & 0.2930 $\pm$ 0.0182 & 1.0843 $\pm$ 0.0328 & 0.3103 $\pm$ 0.0470 & 0.3077 $\pm$ 0.0014 \\
             & opt + smooth prior ($\lambda{=}10^{-3}$) & 0.4232 $\pm$ 0.0059 & 0.2873 $\pm$ 0.0187 & 1.0842 $\pm$ 0.0328 & 0.3074 $\pm$ 0.0474 & 0.3075 $\pm$ 0.0016 \\
\cline{1-7}
\multirow{3}{*}{MTAD-GAT}
             & forward (no opt)                       & 0.2917 $\pm$ 0.0045 & 0.4617 $\pm$ 0.0143 & 1.4727 $\pm$ 0.0705 & 0.3592 $\pm$ 0.0095 & 0.3593 $\pm$ 0.0024 \\
             & opt, $\lambda{=}0$ (no prior)          & 0.2699 $\pm$ 0.0014 & 0.4254 $\pm$ 0.0171 & 1.3859 $\pm$ 0.0842 & 0.3157 $\pm$ 0.0142 & 0.3418 $\pm$ 0.0083 \\
             & opt + smooth prior ($\lambda{=}10^{-3}$) & 0.2701 $\pm$ 0.0014 & 0.4274 $\pm$ 0.0169 & 1.3859 $\pm$ 0.0842 & 0.3157 $\pm$ 0.0141 & 0.3423 $\pm$ 0.0081 \\
\cline{1-7}
\multirow{3}{*}{TCN-AE}
             & forward (no opt)                       & 0.2585 $\pm$ 0.0011 & 0.3856 $\pm$ 0.0272 & 0.9277 $\pm$ 0.0186 & 0.4285 $\pm$ 0.0126 & 0.3292 $\pm$ 0.0044 \\
             & opt, $\lambda{=}0$ (no prior)          & 0.2182 $\pm$ 0.0010 & 0.1940 $\pm$ 0.0052 & 0.9069 $\pm$ 0.0190 & 0.1822 $\pm$ 0.0167 & 0.3121 $\pm$ 0.0008 \\
             & opt + smooth prior ($\lambda{=}10^{-3}$) & 0.2180 $\pm$ 0.0010 & 0.1933 $\pm$ 0.0052 & 0.9068 $\pm$ 0.0190 & 0.1780 $\pm$ 0.0173 & 0.3120 $\pm$ 0.0008 \\
\cline{1-7}
\multirow{3}{*}{TransformerAE}
             & forward (no opt)                       & 0.2575 $\pm$ 0.0039 & 0.6651 $\pm$ 0.1522 & 0.9027 $\pm$ 0.0239 & 0.4165 $\pm$ 0.1061 & 0.3211 $\pm$ 0.0006 \\
             & opt, $\lambda{=}0$ (no prior)          & 0.2074 $\pm$ 0.0011 & 0.3451 $\pm$ 0.1150 & 0.8650 $\pm$ 0.0232 & \underline{0.1624 $\pm$ 0.0199} & 0.3064 $\pm$ 0.0013 \\
             & opt + smooth prior ($\lambda{=}10^{-3}$) & 0.2073 $\pm$ 0.0010 & 0.3367 $\pm$ 0.1167 & 0.8652 $\pm$ 0.0232 & \textbf{0.1606 $\pm$ 0.0194} & 0.3059 $\pm$ 0.0013 \\
\bottomrule
\end{tabular}
}\caption{\textbf{Normal test set evaluation under point missingness (MAE; mean $\pm$ std where applicable).}
MeanFill is a deterministic per-channel training-mean imputer and is reported as a single value. BRITS, FreTS, SAITS, TimesNet, and iTransformer are SOTA baselines evaluated under point missingness. LSTM, MTAD-GAT, TCN-AE, and TransformerAE are frozen reconstructors trained without artificial missingness and evaluated with forward-only inference and test-time input optimization. Best results are marked in bold; second-best results are underlined.}
\label{tab:normal_test_point}
\end{table*}

\begin{table*}[t]
\centering
\small
\resizebox{\textwidth}{!}{\begin{tabular}{ll|ccccc}
\toprule
             & Setting & BeijingAir & ETTh1 & Electricity & ItalyAir & PeMS \\
\midrule
MeanFill
             & train mean & 0.7610 & 0.9757 & 1.8906 & 0.7615 & 0.8097 \\
\midrule
BRITS
             & subseq & \textbf{0.1417 $\pm$ 0.0008} & \underline{0.3876 $\pm$ 0.0223} & 1.2899 $\pm$ 0.0085 & 0.3697 $\pm$ 0.0187 & \underline{0.3208 $\pm$ 0.0014} \\
FreTS
             & subseq & 0.2324 $\pm$ 0.0065 & 0.4546 $\pm$ 0.0281 & 1.0865 $\pm$ 0.0884 & 0.3765 $\pm$ 0.0103 & 0.3980 $\pm$ 0.0090 \\
SAITS
             & subseq & \underline{0.1602 $\pm$ 0.0020} & \textbf{0.3414 $\pm$ 0.0106} & 1.5867 $\pm$ 0.0628 & \textbf{0.2575 $\pm$ 0.0157} & \textbf{0.3184 $\pm$ 0.0011} \\
TimesNet
             & subseq & 0.4642 $\pm$ 0.0074 & 0.9219 $\pm$ 0.0112 & 1.7342 $\pm$ 0.0094 & 0.5939 $\pm$ 0.0070 & 0.3440 $\pm$ 0.0033 \\
iTransformer
             & subseq & 0.6077 $\pm$ 0.0805 & 0.8846 $\pm$ 0.0043 & 1.6973 $\pm$ 0.0693 & 0.5060 $\pm$ 0.0253 & 0.5700 $\pm$ 0.0217 \\
\midrule
\midrule
\multirow{3}{*}{LSTM}
             & forward (no opt)                       & 0.5268 $\pm$ 0.0112 & 0.7869 $\pm$ 0.0026 & 1.1682 $\pm$ 0.0426 & 0.5679 $\pm$ 0.0494 & 0.3715 $\pm$ 0.0054 \\
             & opt, $\lambda{=}0$ (no prior)          & 0.4494 $\pm$ 0.0089 & 0.5533 $\pm$ 0.0251 & 1.1279 $\pm$ 0.0367 & 0.3755 $\pm$ 0.0405 & 0.3583 $\pm$ 0.0024 \\
             & opt + smooth prior ($\lambda{=}10^{-3}$) & 0.4168 $\pm$ 0.0065 & 0.4773 $\pm$ 0.0250 & 1.1279 $\pm$ 0.0367 & 0.3665 $\pm$ 0.0408 & 0.3582 $\pm$ 0.0029 \\
\cline{1-7}
\multirow{3}{*}{MTAD-GAT}
             & forward (no opt)                       & 0.3059 $\pm$ 0.0059 & 0.6623 $\pm$ 0.0165 & 1.5176 $\pm$ 0.0682 & 0.4724 $\pm$ 0.0114 & 0.4166 $\pm$ 0.0031 \\
             & opt, $\lambda{=}0$ (no prior)          & 0.2775 $\pm$ 0.0022 & 0.5202 $\pm$ 0.0343 & 1.4845 $\pm$ 0.0807 & 0.3906 $\pm$ 0.0185 & 0.3953 $\pm$ 0.0097 \\
             & opt + smooth prior ($\lambda{=}10^{-3}$) & 0.2775 $\pm$ 0.0023 & 0.5175 $\pm$ 0.0359 & 1.4845 $\pm$ 0.0807 & 0.3917 $\pm$ 0.0178 & 0.3980 $\pm$ 0.0085 \\
\cline{1-7}
\multirow{3}{*}{TCN-AE}
             & forward (no opt)                       & 0.3207 $\pm$ 0.0018 & 0.7947 $\pm$ 0.0762 & 1.0273 $\pm$ 0.0200 & 0.6806 $\pm$ 0.0196 & 0.3819 $\pm$ 0.0041 \\
             & opt, $\lambda{=}0$ (no prior)          & 0.2672 $\pm$ 0.0015 & 0.5316 $\pm$ 0.0086 & \underline{1.0025 $\pm$ 0.0215} & 0.5988 $\pm$ 0.0388 & 0.3717 $\pm$ 0.0011 \\
             & opt + smooth prior ($\lambda{=}10^{-3}$) & 0.2652 $\pm$ 0.0011 & 0.5313 $\pm$ 0.0086 & \textbf{1.0024 $\pm$ 0.0215} & 0.5441 $\pm$ 0.0374 & 0.3713 $\pm$ 0.0011 \\
\cline{1-7}
\multirow{3}{*}{TransformerAE}
             & forward (no opt)                       & 0.3502 $\pm$ 0.0052 & 0.9488 $\pm$ 0.1760 & 1.0727 $\pm$ 0.0244 & 0.6143 $\pm$ 0.0493 & 0.3738 $\pm$ 0.0013 \\
             & opt, $\lambda{=}0$ (no prior)          & 0.2804 $\pm$ 0.0017 & 0.7832 $\pm$ 0.0633 & 1.0328 $\pm$ 0.0298 & 0.3566 $\pm$ 0.0365 & 0.3638 $\pm$ 0.0011 \\
             & opt + smooth prior ($\lambda{=}10^{-3}$) & 0.2788 $\pm$ 0.0015 & 0.7173 $\pm$ 0.0635 & 1.0331 $\pm$ 0.0298 & \underline{0.3335 $\pm$ 0.0265} & 0.3621 $\pm$ 0.0011 \\
\bottomrule
\end{tabular}
}\caption{\textbf{Normal test set evaluation under subsequence missingness (MAE; mean $\pm$ std where applicable).}
MeanFill is a deterministic per-channel training-mean imputer and is reported as a single value. BRITS, FreTS, SAITS, TimesNet, and iTransformer are SOTA baselines evaluated under subsequence missingness. LSTM, MTAD-GAT, TCN-AE, and TransformerAE are frozen reconstructors trained without artificial missingness and evaluated with forward-only inference and test-time input optimization. Best results are marked in bold; second-best results are underlined.}
\label{tab:normal_test_subseq}
\end{table*}

\begin{table*}[t]
\centering
\small
\resizebox{\textwidth}{!}{\begin{tabular}{ll|ccccc}
\toprule
             & Setting & BeijingAir & ETTh1 & Electricity & ItalyAir & PeMS \\
\midrule
MeanFill
             & train mean & 0.7534 & 0.8154 & 1.8593 & 0.8918 & 0.7972 \\
\midrule
BRITS
             & block & \textbf{0.1474 $\pm$ 0.0012} & 0.4695 $\pm$ 0.0100 & 1.0111 $\pm$ 0.0197 & 0.3635 $\pm$ 0.0027 & 0.3083 $\pm$ 0.0031 \\
FreTS
             & block & 0.2054 $\pm$ 0.0207 & 0.4226 $\pm$ 0.0385 & \underline{0.9126 $\pm$ 0.2514} & 0.4007 $\pm$ 0.0308 & 0.3880 $\pm$ 0.0067 \\
SAITS
             & block & \underline{0.1683 $\pm$ 0.0018} & \textbf{0.3448 $\pm$ 0.0806} & 1.5833 $\pm$ 0.0267 & \textbf{0.2476 $\pm$ 0.0112} & \underline{0.3043 $\pm$ 0.0018} \\
TimesNet
             & block & 0.4482 $\pm$ 0.0081 & 0.7725 $\pm$ 0.0083 & 1.7068 $\pm$ 0.0106 & 0.6722 $\pm$ 0.0077 & 0.3308 $\pm$ 0.0022 \\
iTransformer
             & block & 0.1951 $\pm$ 0.0066 & \underline{0.3939 $\pm$ 0.0086} & \textbf{0.6942 $\pm$ 0.1106} & 0.3394 $\pm$ 0.0403 & \textbf{0.3010 $\pm$ 0.0143} \\
\midrule
\midrule
\multirow{3}{*}{LSTM}
             & forward (no opt)                       & 0.5264 $\pm$ 0.0092 & 0.7706 $\pm$ 0.0034 & 1.1792 $\pm$ 0.0376 & 0.7185 $\pm$ 0.0597 & 0.3571 $\pm$ 0.0058 \\
             & opt, $\lambda{=}0$ (no prior)          & 0.4749 $\pm$ 0.0083 & 0.8890 $\pm$ 0.0486 & 1.1350 $\pm$ 0.0353 & 0.4506 $\pm$ 0.0427 & 0.3435 $\pm$ 0.0028 \\
             & opt + smooth prior ($\lambda{=}10^{-3}$) & 0.4334 $\pm$ 0.0072 & 0.7275 $\pm$ 0.0306 & 1.1349 $\pm$ 0.0354 & 0.4117 $\pm$ 0.0391 & 0.3436 $\pm$ 0.0029 \\
\cline{1-7}
\multirow{3}{*}{MTAD-GAT}
             & forward (no opt)                       & 0.2951 $\pm$ 0.0052 & 0.5144 $\pm$ 0.0097 & 1.4856 $\pm$ 0.0712 & 0.4730 $\pm$ 0.0040 & 0.4035 $\pm$ 0.0027 \\
             & opt, $\lambda{=}0$ (no prior)          & 0.2704 $\pm$ 0.0023 & 0.4815 $\pm$ 0.0143 & 1.4272 $\pm$ 0.0761 & 0.3865 $\pm$ 0.0248 & 0.3807 $\pm$ 0.0102 \\
             & opt + smooth prior ($\lambda{=}10^{-3}$) & 0.2702 $\pm$ 0.0024 & 0.4803 $\pm$ 0.0176 & 1.4272 $\pm$ 0.0761 & 0.3866 $\pm$ 0.0244 & 0.3840 $\pm$ 0.0094 \\
\cline{1-7}
\multirow{3}{*}{TCN-AE}
             & forward (no opt)                       & 0.3471 $\pm$ 0.0013 & 0.7166 $\pm$ 0.0141 & 1.0702 $\pm$ 0.0174 & 0.7647 $\pm$ 0.0159 & 0.3689 $\pm$ 0.0042 \\
             & opt, $\lambda{=}0$ (no prior)          & 0.2899 $\pm$ 0.0026 & 0.5184 $\pm$ 0.0156 & 1.0230 $\pm$ 0.0197 & 0.6337 $\pm$ 0.0392 & 0.3564 $\pm$ 0.0009 \\
             & opt + smooth prior ($\lambda{=}10^{-3}$) & 0.2859 $\pm$ 0.0015 & 0.4837 $\pm$ 0.0079 & 1.0227 $\pm$ 0.0197 & 0.5662 $\pm$ 0.0253 & 0.3559 $\pm$ 0.0009 \\
\cline{1-7}
\multirow{3}{*}{TransformerAE}
             & forward (no opt)                       & 0.3832 $\pm$ 0.0049 & 0.8001 $\pm$ 0.0416 & 1.0665 $\pm$ 0.0235 & 0.6863 $\pm$ 0.0718 & 0.3608 $\pm$ 0.0013 \\
             & opt, $\lambda{=}0$ (no prior)          & 0.2966 $\pm$ 0.0017 & 0.9486 $\pm$ 0.0600 & 0.9956 $\pm$ 0.0242 & 0.3718 $\pm$ 0.0410 & 0.3491 $\pm$ 0.0012 \\
             & opt + smooth prior ($\lambda{=}10^{-3}$) & 0.2930 $\pm$ 0.0018 & 0.7698 $\pm$ 0.0226 & 0.9960 $\pm$ 0.0242 & \underline{0.3253 $\pm$ 0.0281} & 0.3475 $\pm$ 0.0012 \\
\bottomrule
\end{tabular}
}\caption{\textbf{Normal test set evaluation under block missingness (MAE; mean $\pm$ std where applicable).}
MeanFill is a deterministic per-channel training-mean imputer and is reported as a single value. BRITS, FreTS, SAITS, TimesNet, and iTransformer are SOTA baselines evaluated under block missingness. LSTM, MTAD-GAT, TCN-AE, and TransformerAE are frozen reconstructors trained without artificial missingness and evaluated with forward-only inference and test-time input optimization. Best results are marked in bold; second-best results are underlined.}
\label{tab:normal_test_block}
\end{table*}

\subsection{Clock Runtime}
\label{app:clock_runtime}
Table~\ref{tab:runtime_detailed_tto_split} reports wall-clock inference time on the full-channel test split.
The benchmark rows measure one-shot imputation, while the TTO rows separately report optimization without the smoothness prior and optimization with the prior, making the accuracy--runtime tradeoff explicit.
Because the test splits contain different numbers of windows, Table~\ref{tab:runtime_normalized_tto_iterations} additionally normalizes runtime to seconds per 1000 windows and reports the median/p90 stopping iterations for TTO.
The normalized table is generated from the runtime summary CSVs: \texttt{benchmark\_runtime\_summary.csv}, \texttt{autoencoder\_runtime\_summary.csv}, and the corresponding batch-probe metadata in \texttt{batch\_probe\_summary.csv}.

\begin{table}[t]
\centering
\small
\resizebox{\linewidth}{!}{\begin{tabular}{l|ccccc}
\toprule
Model & BeijingAir & ETT h1 & Electricity & ItalyAir & PeMS \\
\midrule
BRITS        & 0.501 & 0.309 & 2.214 & 0.139 & 0.347 \\
FreTS        & 0.056 & 0.022 & 0.788 & 0.051 & 0.102 \\
SAITS        & 0.038 & 0.149 & 0.869 & 0.021 & 0.156 \\
TimesNet     & 0.105 & 0.008 & 0.865 & 0.010 & 0.144 \\
iTransformer & 0.391 & 0.014 & 1.484 & 0.050 & 0.796 \\
\midrule
LSTM + TTO (no prior)               & 10.341 & 2.255 & 5.276   & 4.987 & 2.912 \\
LSTM + TTO (with prior)             & 10.223 & 2.486 & 6.314   & 5.053 & 3.249 \\
MTAD-GAT + TTO (no prior)            & 12.737 & 1.383 & 193.631 & 1.177 & 171.675 \\
MTAD-GAT + TTO (with prior)          & 12.783 & 1.409 & 194.713 & 1.258 & 171.788 \\
TCN-AE + TTO (no prior)              & 12.381 & 3.510 & 10.659  & 6.330 & 3.251 \\
TCN-AE + TTO (with prior)            & 12.641 & 3.554 & 11.631  & 6.423 & 3.442 \\
TransformerAE + TTO (no prior)      & 5.480  & 3.969 & 20.833  & 6.393 & 3.731 \\
TransformerAE + TTO (with prior)    & 5.563  & 4.045 & 21.790  & 6.606 & 4.040 \\
\bottomrule
\end{tabular}
}
\caption{Wall-clock inference time in seconds on the full test set on identical hardware. Benchmark rows report the fastest one-shot runtime across training missingness types (\texttt{point}/\texttt{subseq}/\texttt{block}). TTO rows report the runtime for each optimization variant separately.}
\label{tab:runtime_detailed_tto_split}
\end{table}

\begin{table*}[t]
\centering
\small
\resizebox{\textwidth}{!}{\begin{tabular}{l|ccccc}
\toprule
Model & BeijingAir (N=304) & ETT h1 (N=72) & Electricity (N=303) & ItalyAir (N=155) & PeMS (N=153) \\
\midrule
\multicolumn{6}{l}{\textit{Fastest one-shot benchmark latency (s / 1000 windows)}} \\
BRITS & 1.65 & 4.29 & 7.31 & 0.90 & 2.27 \\
FreTS & 0.18 & 0.31 & 2.60 & 0.33 & 0.67 \\
SAITS & 0.13 & 2.08 & 2.87 & 0.14 & 1.02 \\
TimesNet & 0.34 & 0.11 & 2.85 & 0.07 & 0.94 \\
iTransformer & 1.29 & 0.20 & 4.90 & 0.32 & 5.20 \\
\midrule
\multicolumn{6}{l}{\textit{TTO latency (s / 1000 windows; median/p90 stopping iterations in parentheses)}} \\
LSTM + TTO (no prior) & 34.02 (201/203) & 31.32 (201/328) & 17.41 (201/201) & 32.18 (201/204) & 19.04 (202/203) \\
LSTM + TTO (with prior) & 33.63 (201/203) & 34.53 (201/327) & 20.84 (201/201) & 32.60 (201/203) & 21.24 (201/203) \\
MTAD-GAT + TTO (no prior) & 41.90 (202/204) & 19.22 (206/209) & 639.05 (201/201) & 7.59 (204/208) & 1122.06 (201/201) \\
MTAD-GAT + TTO (with prior) & 42.05 (202/204) & 19.57 (205/208) & 642.62 (201/201) & 8.12 (204/208) & 1122.80 (201/201) \\
TCN-AE + TTO (no prior) & 40.73 (201/203) & 48.75 (208/236) & 35.18 (201/201) & 40.84 (201/208) & 21.25 (201/201) \\
TCN-AE + TTO (with prior) & 41.58 (201/202) & 49.36 (207/235) & 38.39 (201/201) & 41.44 (201/206) & 22.50 (201/201) \\
TransformerAE + TTO (no prior) & 18.03 (201/203) & 55.13 (201/212) & 68.75 (201/201) & 41.25 (201/201) & 24.39 (201/201) \\
TransformerAE + TTO (with prior) & 18.30 (201/202) & 56.18 (201/213) & 71.91 (201/201) & 42.62 (201/201) & 26.41 (201/201) \\
\bottomrule
\end{tabular}
}
\caption{\textbf{Normalized runtime and stopping iterations on the full-channel test split.}
Columns report the number of test windows used for each dataset.
Benchmark entries are seconds per 1000 windows for the fastest one-shot row across point-, subsequence-, and block-trained checkpoints.
TTO entries are seconds per 1000 windows; parentheses report median/p90 stopping iterations across optimized samples.
Measurements use the same hardware, auto-selected batch sizes, warmup runs, and median-of-three timing protocol as Table~\ref{tab:runtime_detailed_tto_split}.}
\label{tab:runtime_normalized_tto_iterations}
\end{table*}

\subsection{Convergence}
\label{appendix-experiments-convergence}
\begin{figure}[h]
    \centering
    \includegraphics[width=\linewidth]{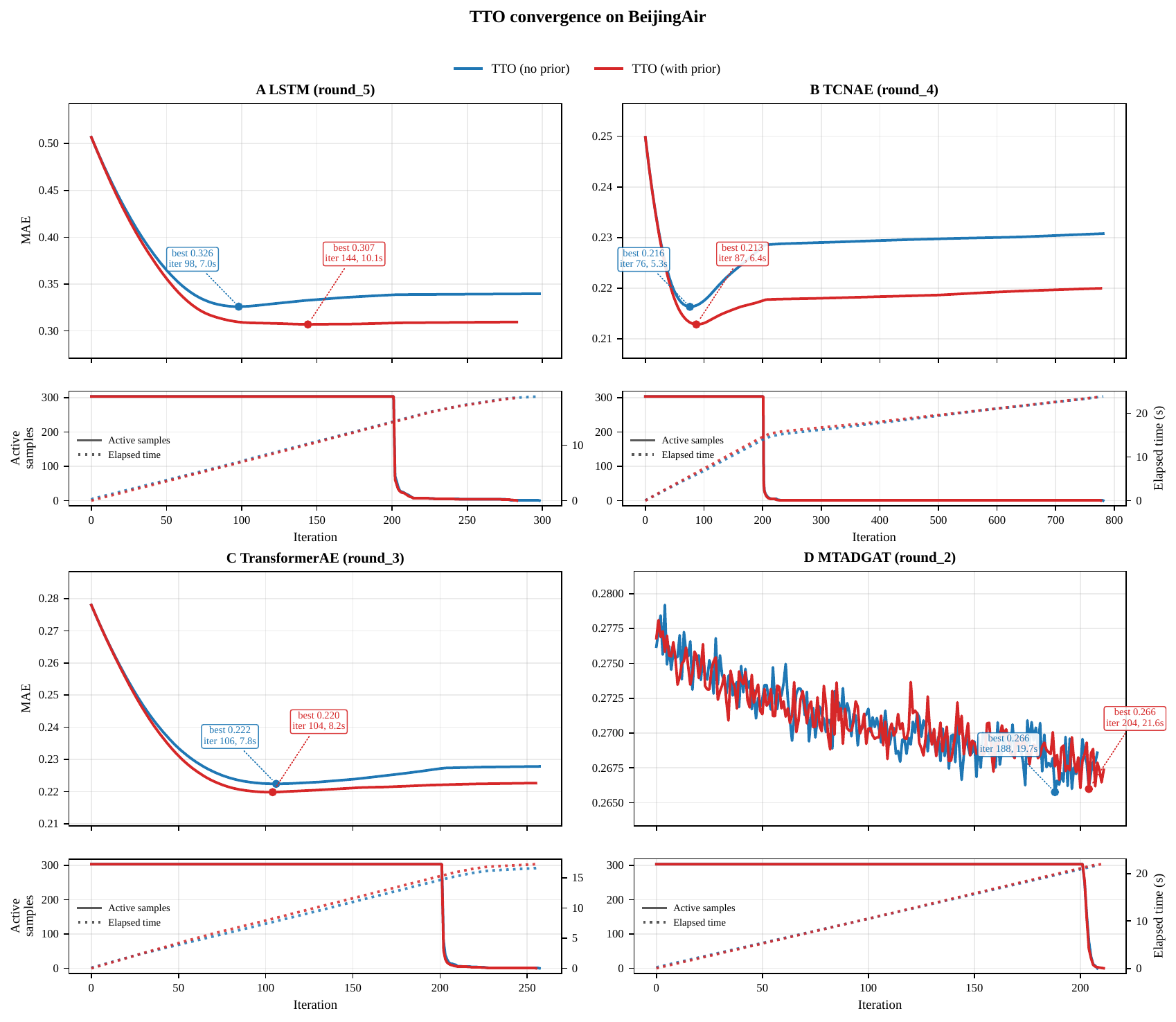}
    \caption{\textbf{Test-Time Optimization convergence on BeijingAir} Convergence behavior of test-time optimization (TTO) for four backbone architectures on the full-channel-missing setting. Each panel reports the tracked MAE over optimization iterations for TTO without prior regularization and TTO with prior regularization. Markers denote the best MAE reached by each variant for diagnostic visualization only, with annotations indicating the corresponding iteration and elapsed time. The lower subplots show the number of active samples during optimization, while dotted curves indicate elapsed wall-clock time.}
    \label{fig:convergence-Beijing}
\end{figure}

\begin{figure}[h]
    \centering
    \includegraphics[width=\linewidth]{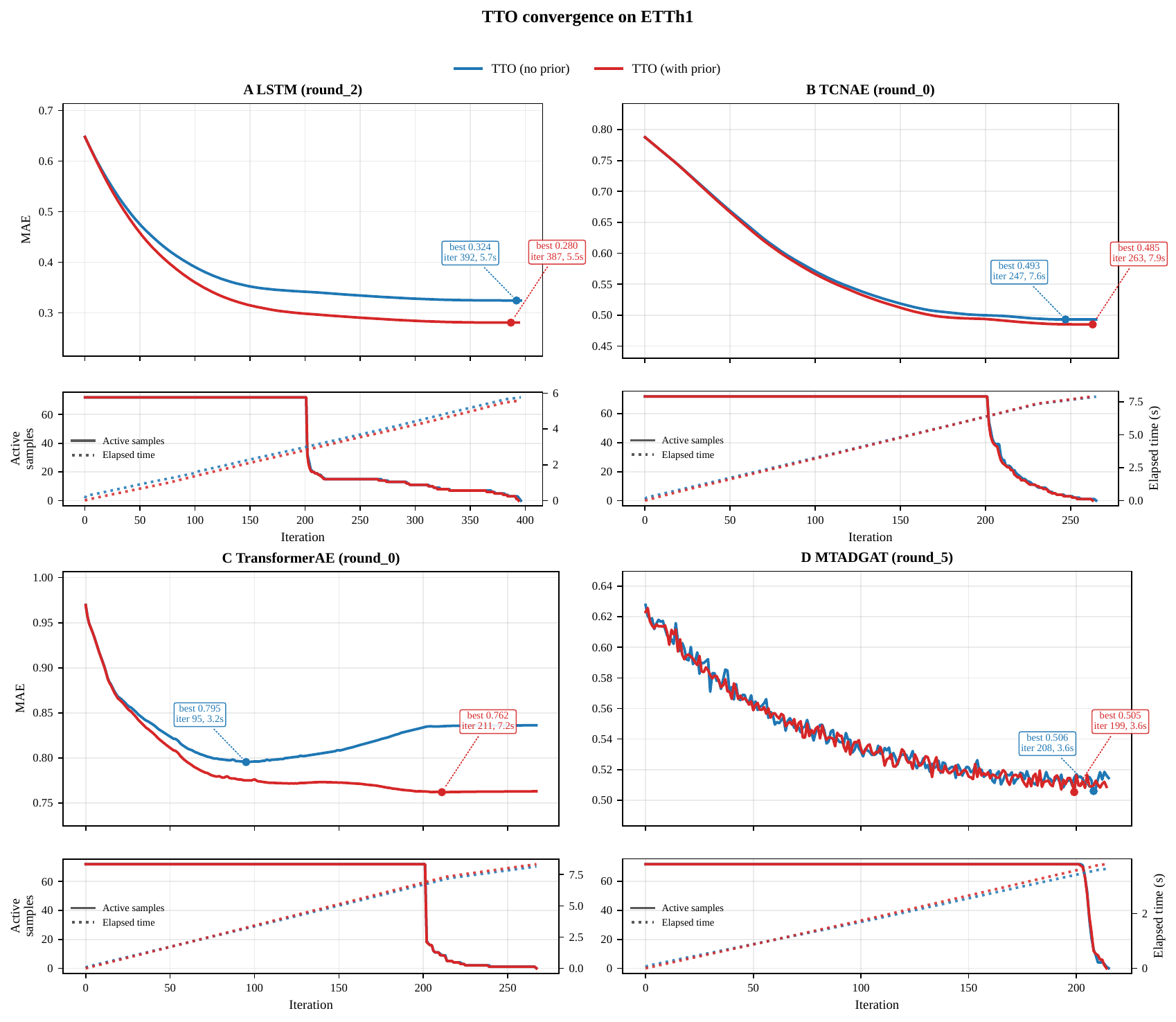}
    \caption{\textbf{Test-Time Optimization convergence on ETTh1} Convergence behavior of test-time optimization (TTO) for four backbone architectures on the full-channel-missing setting. Each panel reports the tracked MAE over optimization iterations for TTO without prior regularization and TTO with prior regularization. Markers denote the best MAE reached by each variant for diagnostic visualization only, with annotations indicating the corresponding iteration and elapsed time. The lower subplots show the number of active samples during optimization, while dotted curves indicate elapsed wall-clock time. For the two backbones LSTM and TransformerAE the prior helps convergence.}
    \label{fig:convergence-ETTh1}
\end{figure}

\begin{figure}[h]
    \centering
    \includegraphics[width=\linewidth]{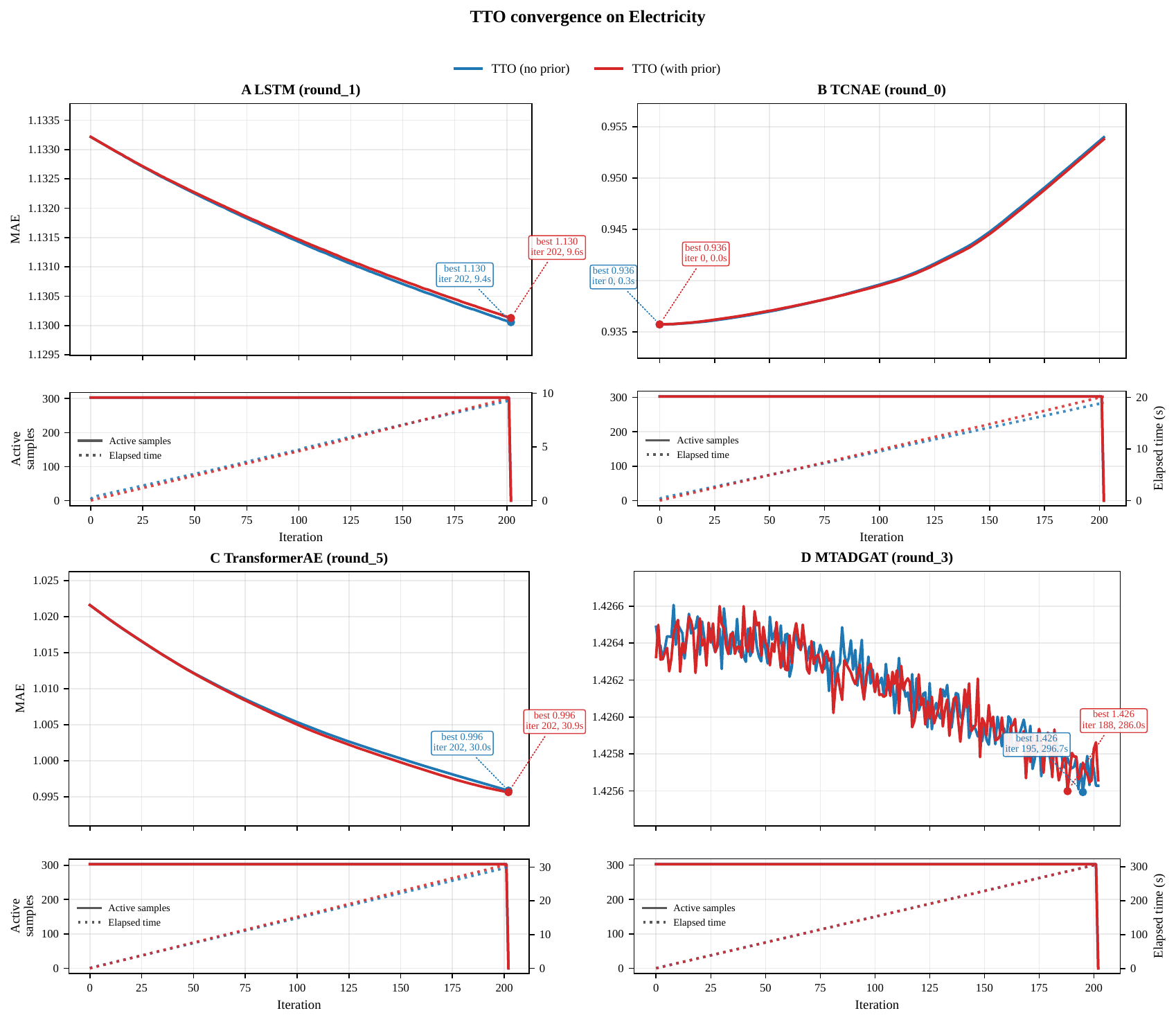}
    \caption{\textbf{Test-Time Optimization convergence on Electricity} Convergence behavior of test-time optimization (TTO) for four backbone architectures on the full-channel-missing setting. Each panel reports the tracked MAE over optimization iterations for TTO without prior regularization and TTO with prior regularization. Markers denote the best MAE reached by each variant for diagnostic visualization only, with annotations indicating the corresponding iteration and elapsed time. The lower subplots show the number of active samples during optimization, while dotted curves indicate elapsed wall-clock time. For the backbone TCN-AE TTO does not improve the imputation performance.}
    \label{fig:convergence-Electricity}
\end{figure}

\begin{figure}[h]
    \centering
    \includegraphics[width=\linewidth]{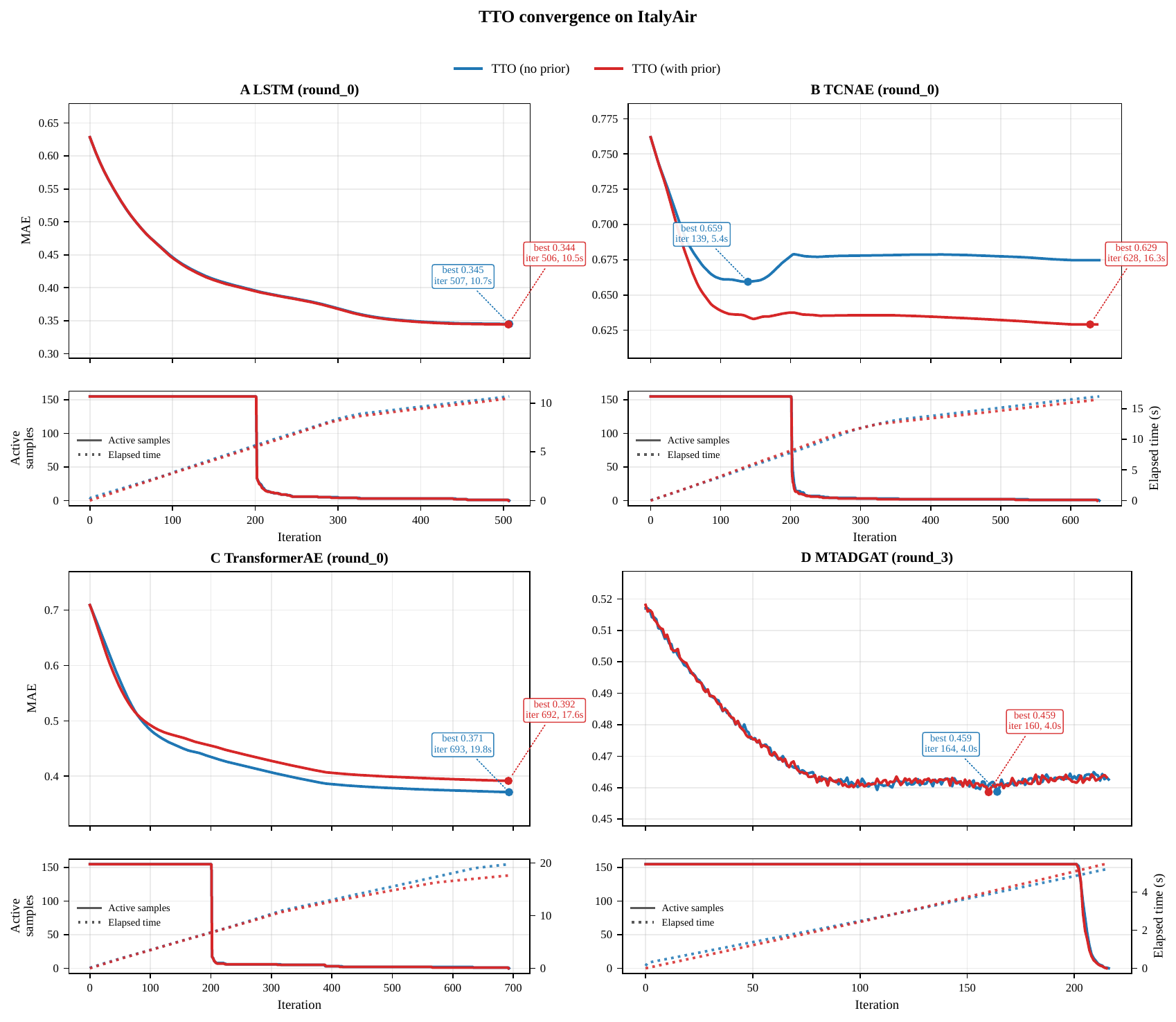}
    \caption{\textbf{Test-Time Optimization convergence on ItalyAir} Convergence behavior of test-time optimization (TTO) for four backbone architectures on the full-channel-missing setting. Each panel reports the tracked MAE over optimization iterations for TTO without prior regularization and TTO with prior regularization. Markers denote the best MAE reached by each variant for diagnostic visualization only, with annotations indicating the corresponding iteration and elapsed time. The lower subplots show the number of active samples during optimization, while dotted curves indicate elapsed wall-clock time. For the backbone TCN-AE the prior helps convergence.}
    \label{fig:convergence-ItalyAir}
\end{figure}

\begin{figure}[h]
    \centering
    \includegraphics[width=\linewidth]{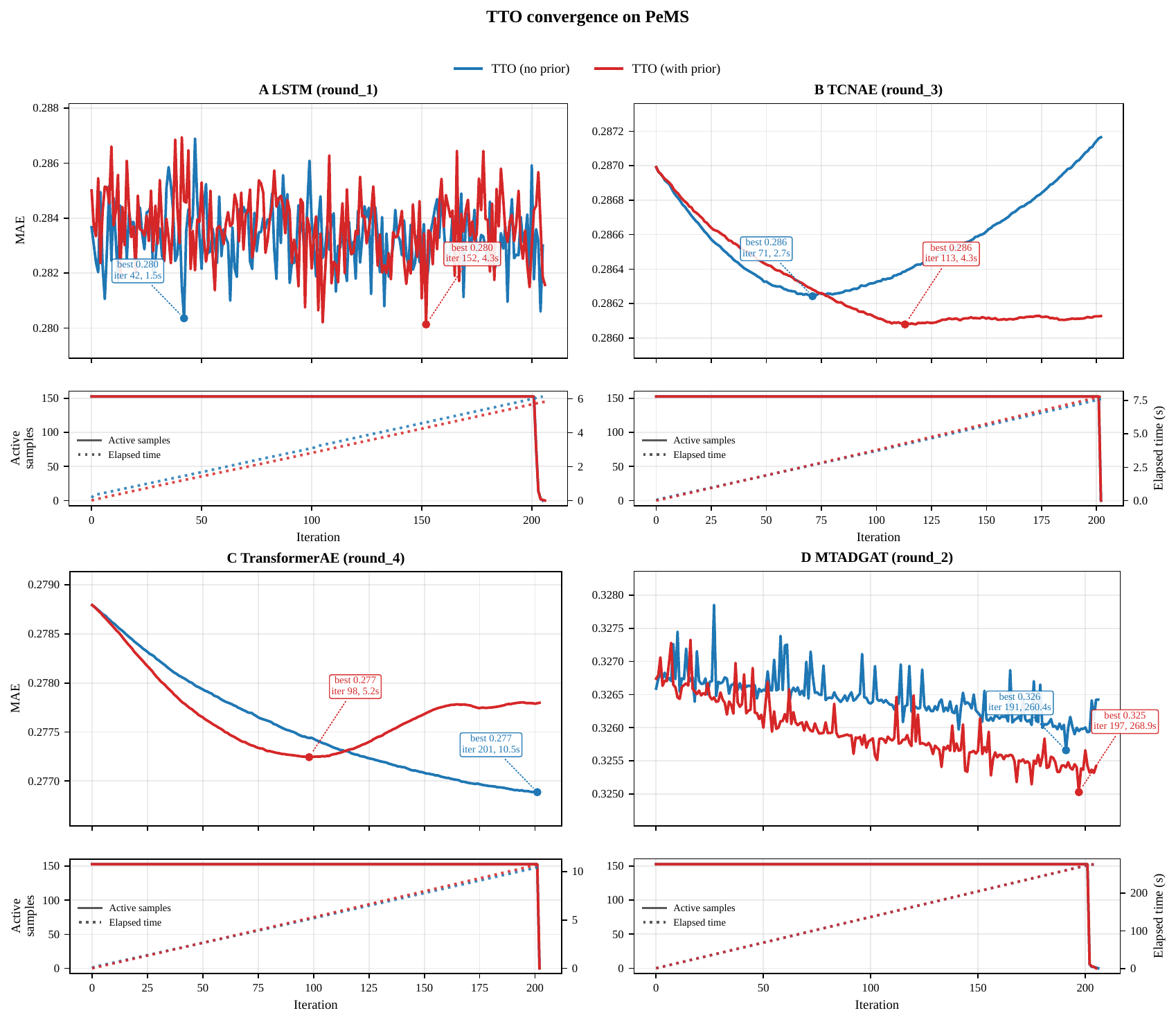}
    \caption{\textbf{Test-Time Optimization convergence on PeMS} Convergence behavior of test-time optimization (TTO) for four backbone architectures on the full-channel-missing setting. Each panel reports the tracked MAE over optimization iterations for TTO without prior regularization and TTO with prior regularization. Markers denote the best MAE reached by each variant for diagnostic visualization only, with annotations indicating the corresponding iteration and elapsed time. The lower subplots show the number of active samples during optimization, while dotted curves indicate elapsed wall-clock time. For the backbone TCN-AE the prior improves convergence, while for the backbone Transformer-AE it does not.}
    \label{fig:convergence-PeMS}
\end{figure}

Figures~\ref{fig:convergence-Beijing}, \ref{fig:convergence-ETTh1}, \ref{fig:convergence-Electricity}, \ref{fig:convergence-ItalyAir}, and \ref{fig:convergence-PeMS} visualize TTO convergence on the full-channel-missing setting for each dataset.
These plots are diagnostic only: they show test MAE over optimization iterations, and the marked best-MAE iterations are not used to select the table results, which use the objective-based stopping rule described in Sec.~\ref{sec:experiments:tti_details}.

\end{document}